\documentclass[10pt,twocolumn,letterpaper]{article}

\usepackage[pagenumbers]{iccv} 

\usepackage{multirow}
\usepackage{pifont}
\usepackage{bm}

\newcommand{\gl}[1]{``\textit{#1}''}
\newcommand{\ci}[3]{#1\,{\scriptsize[#2,\,#3]}}

\newcommand{\fitwidth}[1]{\resizebox{\ifdim\width>\linewidth\linewidth\else\width\fi}{!}{#1}}

\definecolor{iccvblue}{rgb}{0.21,0.49,0.74}
\usepackage[pagebackref,breaklinks,colorlinks,allcolors=iccvblue]{hyperref}

\def\paperID{*****} 
\def\confName{ICCV}
\def\confYear{2025}

\title{6px Is Worth a Chinese Character: \\Context Learning for Visually Ambiguous Symbolic Systems}

\author{Yifan Liu\\
Stanford University\\
California, U.S.A. \\
{\tt\small lyf@stanford.edu}} 

\begin{document}
\maketitle

\newcommand{\model}{Phrolova-t\xspace}      
\newcommand{\bench}{TinyHanzi\xspace}       
\newcommand{\cutnet}{CutNet\xspace}         
\newcommand{\clsnet}{CodebookNet\xspace}        
\newcommand{\lpd}{LPD\xspace}               
\newcommand{\lpdlong}{lattice product decoding\xspace}
\newcommand{\gbank}{GlyphBank\xspace}       
\newcommand{\hilo}{H2L\xspace}    

\begin{abstract}
Prior work on extremely low-resolution text targets English and other small alphabets, which does not extend to
Chinese, whose thousands of characters each pack many strokes into a square cell. We study OCR on Chinese text only 5--7
pixels tall and build \bench, a benchmark of
eight sub-benches from six games where every video keeps native-resolution frames next to the shrunk targets. On it, the
best vision-language model reads 47\% of the characters at 5\,px, and super-resolution makes every reader worse. We
attribute this to treating the task as open-ended vision: a $6{\times}6$ RGB patch has $2^{864}$ values, but the answer
is one of about 3,500 characters. We introduce \model, which classifies characters into this codebook and
decodes with two layers of context: within a sentence, it multiplies the classifier's and a language model's
distributions over the candidate lattice instead of post-processing the top-1 string; within a video, high-resolution glyphs teach the low-resolution reader.
\model reads 84.2\% of the characters, 11.3 points above a Qwen3.5-9B LoRA trained on our data (23 at
5\,px), and is on par with strong VLMs on established benchmarks it never trained on. 
The benchmark will be released upon copyright authorization; code is at {\fontsize{9.5}{11}\selectfont\textup{\url{https://github.com/liuyifan22/phrolova_t}}}.
\end{abstract}

\vspace{-0.2em}

\begin{figure}[t]
  \centering
  \includegraphics[width=\columnwidth]{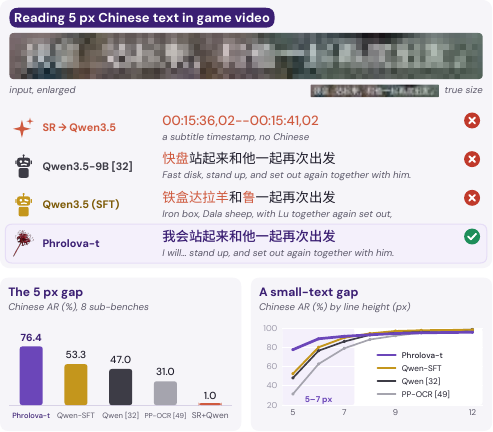}
  \caption{\textbf{Reading 5\,px Chinese.} Top: a real 5\,px line from \textit{Honkai: Star Rail}, enlarged and at true
  size, and what four readers make of it (Red: wrong characters; Grey: literal English of each reading).
  Super-resolution followed by Qwen3.5-9B reads a subtitle timestamp, Qwen3.5-9B and Qwen3.5-9B fine-tuned on our data
  (SFT) invent words, and \model reads it exactly. Bottom left: Chinese accuracy rate (AR) at 5\,px on all eight sub-benches
  of \bench. Bottom right: AR against line height on the same 1,732 lines; every reader is near 99\% at native
  resolution. Game frames: research use only, no commercial use.}
  \label{fig:teaser}
\end{figure}

\section{Introduction}
\label{sec:intro}

Text in videos, such as the dialogue, quests, names and instructions of games, live streams and screen recordings,
carries much of their content, and reading it enables downstream applications such as video search, translation and
indexing. Yet such text is often tiny: a phone recording of a tablet game, a 360p stream, or blurry recordings from ancient video games can
leave a line of Chinese only 5--7 pixels tall (\cref{fig:teaser}). At this size, a character with fifteen strokes has
more strokes than its image has pixel rows, and to an unprepared reader the whole line becomes a smear.

However, reading such text is particularly challenging for Chinese characters. Unlike an alphabet of a few dozen letters,
everyday Chinese uses about 3,500 characters~\cite{gscc2013}, often a dense arrangement of strokes in a square cell, so
that many characters differ by a single stroke. Existing work on ultra-low-resolution
OCR~\cite{jacobs2005lowres,wachenfeld2006screen,gilbey2021ultralow,hill2016mosaic,lee2026unreadable} focuses on English
or other small alphabets, which does not extend to such a complex symbolic system. Video OCR has treated Chinese
characters below 10\,px as unrecognizable~\cite{wang2019ultralow}, and text super-resolution typically starts from
inputs of about 16\,px~\cite{wang2020textzoom,ma2023realce}. Current readers also fail in this regime: on our benchmark,
the best off-the-shelf vision-language model (VLM) reads only 47\% of the Chinese characters at 5\,px, although it reads
99\% of the same lines at native resolution. Moreover, super-resolution (SR) does not help. SR before OCR lowers every
reader we tried (e.g., Qwen3.5-9B from 65.9\% to 20.4\%), since it invents plausible strokes that spell other characters,
or none at all (\cref{fig:teaser}, top).

A human reader who knows the game, however, can often still guess such a line. We do not resolve every stroke;
instead, we know which characters are plausible, what the sentence is likely to say, and how the same words looked when
they appeared large in a title or a menu. This observation suggests that reading tiny text does not necessarily require
recovering the image---only matching the pixels against a small set of known patterns and letting context decide.
Indeed, a $6{\times}6$ RGB patch can take $(256^3)^{36} = 2^{864} \approx 10^{260}$ values, whereas the character it
shows is one of about 3,500, drawn in a handful of fonts and degraded by a known channel (area downsampling and
compression). Reading tiny Chinese is therefore not open-ended visual recognition but \textit{closed-set decoding}.
Inspired by this, we ask: can a reader decode tiny text by comparing its pixels with a finite codebook and resolving
the remaining ambiguity with context, rather than by trying to see more clearly?

We propose \model, a lightweight yet effective framework that follows this principle (\cref{fig:method}). Instead of
generating open text from pixels, \model classifies each character into the finite codebook, keeps the whole
distribution over candidates, and lets two layers of context choose among the characters the pixels allow. The first
layer enforces \textit{coherence within a sentence}: the characters must form plausible Chinese. The second enforces
\textit{coherence within a video} or even a mixture of visual materials: the same characters in the same fonts also appear large in titles, menus, or other
higher resolution videos, where they are easy to read, so that high resolution can teach low.

Specifically, given a line, \cutnet first cuts it into character pieces, and \clsnet classifies each piece into a
distribution over the codebook (\cref{sec:likelihood}). Context layer~1, \lpdlong (\lpd), multiplies this distribution
with a language model's distribution over the lattice of candidates, a noisy-channel product rather than a post-processing correction
of the top-1 string (\cref{sec:layer1}). Context layer~2, \gbank, turns the high-resolution lines of the same video into
a library of glyphs degraded to the target scale and matches them against the tiny pieces at test time
(\cref{sec:layer2}). At training time, \hilo uses the same kind of lines in unlabelled videos to self-label
low-resolution crops in the game's own fonts, without any hand labels.

To evaluate this regime, we build \bench, a benchmark of eight sub-benches from six games with 6,126 lines and 52,236
Chinese characters at 5, 6 and 7\,px, where every video keeps native-resolution context frames next to the shrunk
targets. With only small trainable networks and a frozen language model, \model reads 84.2\% of the Chinese
characters, outperforming Qwen3.5-9B fine-tuned with LoRA on our data by 11.3 points and the best of ten LLM-assisted
OCR pipelines by 14.2 points. Notably, the advantage is a small-text advantage: on the lines rebuilt at every
height, \model leads by 25 points at 5\,px and by 1.5 at 7\,px, and its lead grows with character complexity, from
+6.7 points for characters of 1--5 strokes to +15.7 for those of 15 or more. On established Chinese benchmarks that
\model never trained on, shrunk to 5--7\,px, it remains on par with the strongest VLMs overall and ahead at 5\,px.
These results demonstrate that, for tiny text, the right prior is worth more than more pixels. Our contributions are
summarized as follows:
\begin{itemize}[leftmargin=*,itemsep=1pt,topsep=2pt]
  \item We introduce the view of \textbf{tiny-text OCR as pattern recognition}: reading tiny Chinese is closed-set
  decoding against a finite codebook through a known channel. Our analysis supports this view, showing that about
  90\% of the common characters have a look-alike at 5\,px, that accuracy collapses below 8\,px, and that
  super-resolution cannot recover the text.
  \item We present \textbf{\bench}, to our knowledge the first benchmark of Chinese text at a 5--7\,px character
  height, which mixes shrunk target lines with native-resolution context lines in each video and comes with a
  resolution sweep from 5 to 12\,px and a real 360p twin that matches our simulation ($r = 0.994$).
  \item We show that, for tiny text, a language model should \textbf{decode, not correct}. \lpd multiplies the
  classifier's distribution, divided by its training prior, with a frozen LLM's distribution over the candidate
  lattice, which is the exact noisy-channel posterior and leaves no fusion weight to tune. Unlike LLM post-correction,
  which sees only the top-1 string and often lowers accuracy, the product lets the language model choose only among the
  characters the pixels allow. It is the largest single gain in \model (+8.8 points) and needs no in-domain data.
  \item We propose \textbf{\model}, which pairs \lpd for coherence within a sentence with \gbank and \hilo for
  coherence within a video, letting the high-resolution text of the same domain teach the low-resolution reader.
  \item We demonstrate state-of-the-art accuracy on \bench, with the largest gains at 5\,px and on complex characters,
  and strong generalization to established Chinese benchmarks that \model never trained on.
\end{itemize}

We hope this study fuels research on multimodal learning for visually ambiguous symbolic systems, where many classes,
fine detail and tiny renderings meet, and motivates further exploration of human-inspired priors that let context
compensate for missing pixels.

\begin{figure*}[t]
  \centering
  \includegraphics[width=\textwidth]{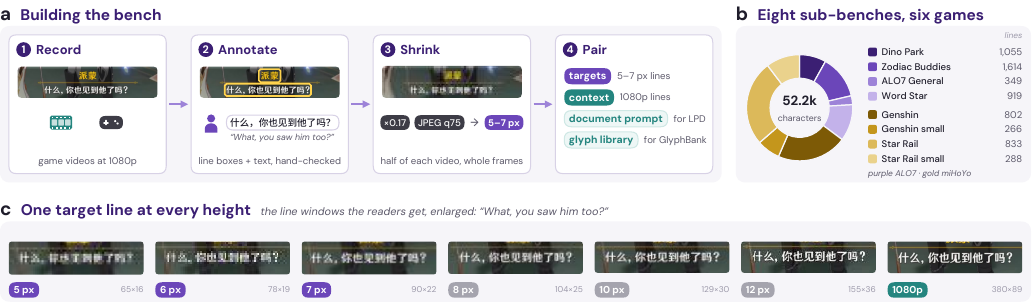}
  \caption{\textbf{The \bench benchmark.} (a) Construction on a real frame: videos are recorded, every line is
  annotated at native resolution, half of the frames of each video are shrunk so that their lines are 5--7\,px tall,
  and the other half stay at native resolution as the targets' context. (b) The eight sub-benches. (c) One line at every
  height from 5 to 12\,px and at 1080p. Game frames: research use only, no commercial use.}
  \label{fig:bench}
\end{figure*}

\section{Related Work}
\label{sec:related}

\noindent\textbf{Chinese text recognition.} Benchmarks such as ReCTS~\cite{zhang2019rects}, CTW~\cite{yuan2019ctw} and the
unified BCTR~\cite{chen2021bctr} cover scene, web, document and handwritten text at ordinary sizes. Recognizers range
from CTC models~\cite{shi2017crnn,du2020ppocr} to methods that exploit radicals and strokes for rare
characters~\cite{zhang2018radical,chen2021stroke}. Before deep learning, handwritten Chinese text was read by
over-segmenting a line, classifying each piece and searching a candidate lattice with a language
model~\cite{wang2012hctr}. Our decoder revives this design with a modern LLM as the prior. None of these works studies
lines a few pixels tall.

\noindent\textbf{Text-aware super-resolution and tiny text.} Text SR restores low-resolution text images before OCR, from
TextZoom and TSRN~\cite{wang2020textzoom} to text-prior and attention models~\cite{ma2023tpgsr,ma2022tatt}, structure
priors for Chinese~\cite{li2023marconet} and diffusion~\cite{zhang2024difftsr}. Real-CE~\cite{ma2023realce} pairs real
low- and high-resolution Chinese photos. Their inputs are about 16\,px or larger, and they output pixels, which at
5\,px are hallucinated (\cref{sec:sr}). Work that reads text below that size uses small alphabets and simple characters: screen-rendered
Latin text~\cite{wachenfeld2006screen}, low-resolution English documents~\cite{jacobs2005lowres,gilbey2021ultralow},
mosaic-redacted text~\cite{hill2016mosaic} and alphanumerics in degraded photos~\cite{lee2026unreadable}. With a few
dozen classes these settings are far easier than a 3,500-class script whose characters fill a square cell with strokes.
Chinese below 10\,px has been considered unreadable~\cite{wang2019ultralow}.

\noindent\textbf{OCR with VLMs and LLMs.} OCR-specialised and general VLMs~\cite{bai2025qwen3vl,cui2025paddleocrvl,%
wei2024got,wei2025deepseekocr,hunyuan2025ocr} now read ordinary Chinese text almost perfectly. LLMs are also used to
correct OCR output, by prompting with the draft and its context~\cite{bourne2024clocrc,kanerva2025nofreelunch}, with the
image as well~\cite{greif2025multimodal}, or through a noisy channel of phonetic and glyph
confusions~\cite{zhou2024simplecsc}. These correctors see only the top-1 string. Combining a recognizer with a
language model inside the search is older: hybrid speech recognizers divide network posteriors by class
priors~\cite{bourlard1994hybrid}, shallow fusion adds an external language model's score during beam
search~\cite{gulcehre2015shallow}, and recent work fuses LLMs with generative ASR and OCR models at the byte
level~\cite{hsu2025gfd}. \lpd applies this rule to the per-character distribution of a closed-set classifier, at the
scale where the pixels alone cannot decide, with a token-level search that lets a subword LLM score a character
lattice; we show that this product, not string correction, closes the gap (\cref{sec:exp-main}).

\section{Analysis: Why 5\,px Chinese Is Hard}
\label{sec:analysis}

To understand why current readers fail on tiny Chinese, we first build a benchmark for this regime
(\cref{sec:bench}). We then use it to show where accuracy collapses (\cref{sec:cliff}), why super-resolution does not
help (\cref{sec:sr}), and what kind of problem remains (\cref{sec:matching}).

\begin{figure*}[t]
  \centering
  \includegraphics[width=\textwidth]{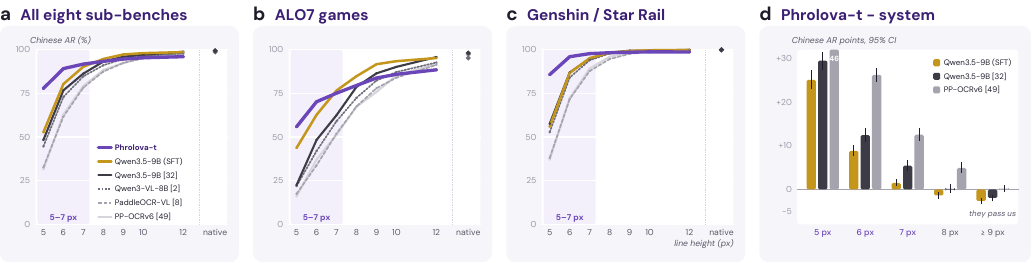}
  \caption{\textbf{The resolution cliff.} (a--c) Chinese AR against line height on the same 1,732 lines, rebuilt at
  every height from 5 to 12\,px, for all eight sub-benches, the ALO7 games and the miHoYo games. Diamonds: each system
  reading the native-resolution line. (d) \model minus each system by height, with 95\% intervals. Existing readers lose
  half of the text between 8 and 5\,px; \model loses much less.}
  \label{fig:cliff}
\end{figure*}

\subsection{The \bench benchmark}
\label{sec:bench}

\noindent\textbf{Sources.} \bench has eight sub-benches from six games (\cref{fig:bench}b). Four come from children's
learning games by ALO7: Dino Park, Zodiac Buddies, ALO7 General and Word Star. Four come from the story videos of two
commercial games, \textit{Genshin Impact} and \textit{Honkai: Star Rail}, each split into a large sub-bench (three
consecutive videos) and a small one (one video). The ALO7 games use rounded, outlined and coloured display fonts on busy
backgrounds; the miHoYo games show long subtitles in a clean sans font. Together they hold 696 frames, 6,126 target
lines with Chinese text and 52,236 Chinese characters.

\noindent\textbf{Mixed resolution in each video.} A frame is turned into a benchmark frame by shrinking it until its median
line is 5, 6 or 7\,px tall (area resampling), followed by JPEG compression (quality 75). Lines of at most 8.5\,px are the
targets. The key design choice is that each video keeps both kinds of text: half of its frames are shrunk and the other
half stay at native resolution, then the halves swap, so every frame is a target once (cross-fitting). Dino Park and
Zodiac Buddies use a single split instead (15 of 32 and 53 of 106 frames are targets, the rest only context). A reader may
use the native-resolution half as context, which is what videos offer in practice: titles, menus and cut scenes in
the same fonts, large enough to read.

\noindent\textbf{Ground truth and realism.} Dino Park, Zodiac Buddies and ALO7 General are transcribed by hand. Word Star
and the miHoYo sub-benches are drafted by PP-OCR at native resolution and checked by hand. To test the simulation, we
also read the miHoYo episodes from a real 640$\times$360 stream of the same videos. Over 27 systems, accuracy on the
real stream and on our simulation of it correlates at $r = 0.994$, and the simulation is harder by a median of 0.85
points. Zodiac Buddies informed development decisions and serves as a dev bench; we report results with and without
it. Construction details and statistics are in \cref{app:bench}.

\noindent\textbf{Metric.} We score Chinese characters only (U+4E00--U+9FFF), since Latin letters, digits and punctuation are
not what makes the regime hard. The primary metric is the Chinese accuracy rate $\mathrm{AR} = 1-\mathrm{CER}$, where a
wrong, missing or added character each costs one, so AR can be negative. We also report exact line accuracy (LA).
Intervals are 95\% paired bootstrap intervals over frames.

\subsection{The resolution cliff}
\label{sec:cliff}

To locate where accuracy collapses, we also rebuild all eight sub-benches at 8, 9, 10 and 12\,px and score the 1,732
lines that are targets at every height (\cref{fig:cliff}). At native resolution every strong reader is near 99\%, so neither the
fonts nor the vocabulary make \bench hard---the challenge is only resolution. Accuracy stays high down to about 9\,px and then falls off a cliff: from
8 to 5\,px, Qwen3.5-9B drops from 93.1\% to 48.3\% and PP-OCRv6 from 88.3\% to 31.6\%. The cliff is steepest on the
ALO7 games, whose display fonts are hardest, but it is just as sharp on the clean miHoYo subtitles, where Qwen3.5-9B
falls from 98.3\% to 57.6\%. \model flattens it: it reads 77.8\% at 5\,px, 25 points more than any baseline (\cref{fig:cliff}d). These results
suggest that five to seven pixels is a distinct regime, which is the focus of this paper.

\subsection{Why super-resolution does not solve it}
\label{sec:sr}

\begin{figure}[h]
  \centering
  \includegraphics[width=\columnwidth]{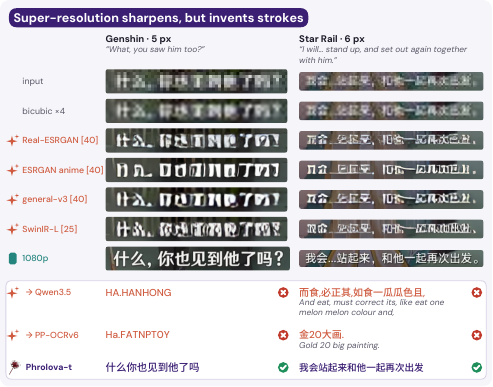}
  \caption{\textbf{Super-resolution sharpens, but invents strokes.} Two real lines (Genshin 5\,px, Star Rail 6\,px)
  through bicubic upscaling and four SR models, with the 1080p line for reference. Below: what Real-ESRGAN followed by
  Qwen3.5-9B or PP-OCRv6 reads, and what \model reads from the raw pixels. Game frames: research use only, no
  commercial use.}
  \label{fig:sr}
\end{figure}

\begin{figure*}[t]
  \centering
  \includegraphics[width=\textwidth]{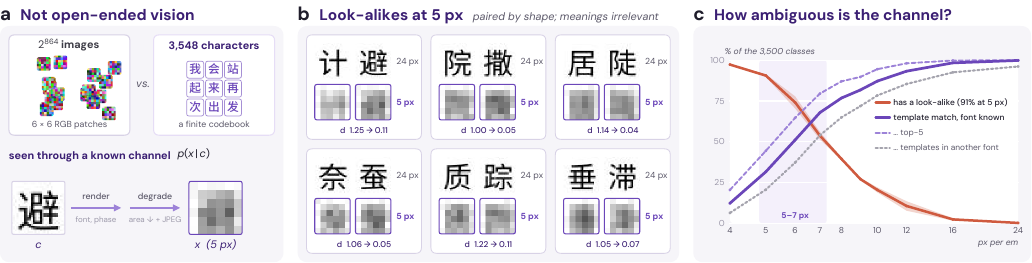}
  \caption{\textbf{Tiny Chinese is a finite codebook seen through a known channel.} (a) The space of $6{\times}6$ RGB
  patches against the codebook of common characters, and the channel $p(x \mid c)$ that renders a character $c$ and
  degrades it to a 5\,px image $x$. (b) Six real look-alike pairs, distinct at 24\,px and nearly identical at 5\,px
  ($d = 2 - 2\rho$, with $\rho$ the correlation of the two mean-removed renders, at 12\,px and at 5\,px). (c) Share of the 3,500 common
  characters with a look-alike within rendering noise, and learning-free template matching, by height.}
  \label{fig:analysis}
\end{figure*}

A natural solution is to super-resolve the line before reading it. However, this fails. As shown in \cref{fig:sr}, we
pass two lines through four SR models: the output is sharp, but its strokes are invented, and different models invent
different ones. Read after
Real-ESRGAN~\cite{wang2021realesrgan}, the 5\,px Genshin line becomes Latin letters (\textit{HA.HANHONG}), and the
6\,px Star Rail line becomes fluent but unrelated classical Chinese (\gl{and eat, must correct its, like eat one
melon}). Over the whole benchmark, Real-ESRGAN before OCR lowers every reader: Qwen3.5-9B from 65.9\% to 20.4\%,
PP-OCRv6 from 54.1\% to 17.0\% and PaddleOCR-VL from 53.8\% to 14.3\%, and at 5\,px it leaves every reader below 3\%
(\cref{app:more}). On naturally blurry ALO7 phone recordings, text-specific SR methods behave the same way: they write
clean, font-like characters that are often confidently wrong (\cref{app:blurry}).

We attribute this failure to the objective of SR, which maximizes how plausible the image looks rather than whether
the characters are right. At
5\,px the pixels are consistent with many characters, and SR picks one by its image prior, often a character-like
shape that is no character at all. This observation suggests that a reader should instead keep the candidates the
pixels allow and choose among them with priors about the text. A deeper reason is that OCR is a classification problem in a very limited search space, which should be much lower in complexity than the vision task of high-resolution reconstruction (\cref{sec:matching}). This paper argues that training SR for text is not a promising direction.

\subsection{OCR: pattern matching, not open-ended vision}
\label{sec:matching}

\noindent\textbf{Counting.} A $6{\times}6$ RGB patch can take $2^{864} \approx 10^{260}$ values. The character it shows is one
of about 3,500 common characters (Level~1 of the Table of General Standard Chinese Characters~\cite{gscc2013}), and
within one game each character is drawn in a few fonts. What reaches the camera is a rendered template passed through
a known channel: sub-pixel placement, area downsampling and compression. Recognition is therefore decoding a finite set
of templates through that channel (\cref{fig:analysis}a), not open-ended visual recognition.

\noindent\textbf{Ambiguity of the channel.} To measure how ambiguous the channel is, we render the 3,500 characters in four fonts (sans regular and bold, rounded,
serif) through the benchmark's degradation at 4--24\,px and measure, without any learning, how often two characters
fall within rendering noise of each other (\cref{fig:analysis}c, \cref{tab:ambiguity} in the appendix). At 5\,px,
90--91\% of the characters have such a look-alike; at 8\,px still 39--40\%; at 16\,px only 2--3\%. The pairs are often
very different characters: the characters for \gl{count} (4 strokes) and \gl{avoid} (16 strokes) become nearly the same
blob (\cref{fig:analysis}b).
The pixels alone do not decide the character.

\noindent\textbf{The truth is usually among a few candidates.} Nevertheless, the pixels narrow the choice down. Learning-free nearest-template matching with the font known reads
28--35\% of the characters at 5\,px, and its top five contain the truth for 41--48\%; at 8\,px, top-1 is already
74--80\%. Knowing the style of the font matters: templates in a font of another style lose 14 points at 6\,px
(51.1\% to 37.2\%), almost all of it between serif and sans. Complex characters suffer most: at 5\,px template
matching reads 25\% of the characters with 15 or more strokes against 44\% of those with 1--5.

\noindent\textbf{Consequences for the design.} Together, these findings motivate four design choices, each of which
becomes a component of \model:
\begin{enumerate*}[label=(\roman*)]
  \item Classify into the finite codebook rather than generate open text.
  \item Keep the full distribution over candidates, not only the top-1 guess.
  \item Combine it with a prior over text by multiplication (context layer~1).
  \item Use same-font, high-resolution examples from the same video as templates (context layer~2).
\end{enumerate*}

\section{Method}
\label{sec:method}

\begin{figure*}[t]
  \centering
  \includegraphics[width=\textwidth]{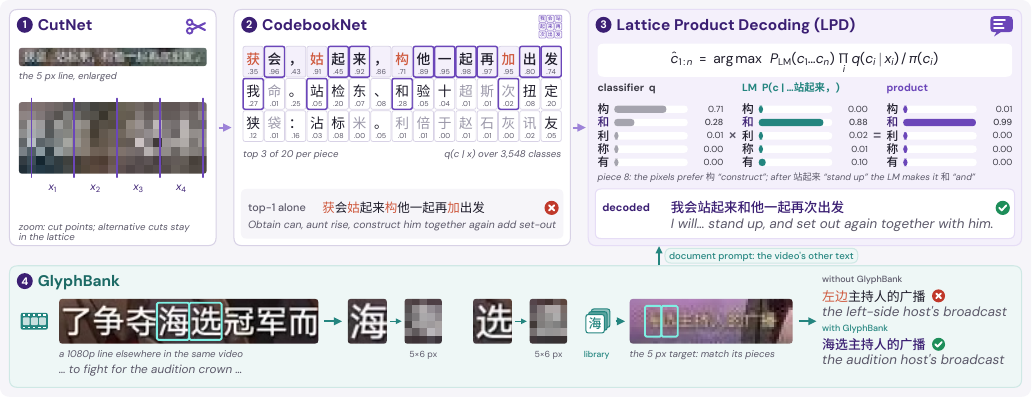}
  \caption{\textbf{\model on a real 5\,px line} (the line of \cref{fig:teaser}). 1)~\cutnet cuts the line into
  pieces; alternative cuts stay in the lattice. 2)~\clsnet classifies each piece into a distribution over the
  codebook (top 3 of the 20 kept candidates shown, boxes: the truth); read alone, its top-1 choices get four
  characters wrong. 3)~\lpd: at piece 8 the pixels prefer the character for \gl{construct}, the language model
  given the decoded prefix prefers the one for \gl{and}, and their product picks \gl{and}; the decoded line is exact.
  4)~\gbank: glyphs cut from a 1080p line of the same video are downsampled to the target's scale and matched
  against the 5\,px pieces, which turns \gl{left-side} into the correct \gl{audition}. The video's other text also
  serves as a document prompt. Game frames: research use only, no
  commercial use.}
  \label{fig:method}
\end{figure*}

\noindent\textbf{Task definition.} We aim to read Chinese text lines that are only 5--7\,px tall. Given a line image
$X$ cropped from a video frame, together with the native-resolution text of the other frames of the same video, the
objective is to output the string of characters $c_{1:n}$ shown in $X$, where each character belongs to a finite
codebook $\mathcal{C}$ of 3,500 common characters and a few symbols.

\noindent\textbf{Framework.} As shown in \cref{fig:method}, \model reads a line in four steps. \cutnet first cuts the
line into character pieces $x_{1:n}$, and \clsnet classifies every piece into a distribution over $\mathcal{C}$
(\cref{sec:likelihood}). The resulting lattice of candidates is then decoded with two layers of context: \lpd enforces
coherence within the sentence with a language model (\cref{sec:layer1}), and \gbank enforces coherence within the video
with glyphs cut from its high-resolution lines (\cref{sec:layer2}). In other words, the first two steps estimate how
well each character explains each piece, while the context layers decide among the characters the pixels allow.

\subsection{Character likelihoods}
\label{sec:likelihood}

\noindent\textbf{Cutting.} Since Chinese characters and full-width punctuation each occupy a square cell, we cut a line
into pieces of about one em. Specifically, \cutnet is a small convolutional network (1.85M parameters) that reads the
line resampled to a fixed 15\,px per em and predicts, for every column, the probability of a cut and of a character
centre. Peaks of the cut probability give the pieces. Because a detector's box is never exact at 5\,px, a companion
network of the same design (1.28M parameters) first refines the line box, and pieces that the classifier labels \texttt{<junk>} are dropped.

\noindent\textbf{Classifying into the codebook.} \clsnet is a ResNet-18~\cite{he2016resnet} that maps a $32{\times}32$ crop
of 1.3\,em around a piece to a softmax over 3,548 classes: the 3,500 common characters, 47 symbols (digits, Latin
letters, punctuation) and \texttt{<junk>}. It is trained on 18.3M crops of synthetic text, rendered in 31 open fonts
on game-like backgrounds and degraded to 5--20\,px by random sub-pixel placement, blur, resampling kernels, chroma
subsampling, JPEG and noise. Instead of keeping only the top-1 label, we keep its full output $q_i(c) \approx p(c \mid x_i)$
for every piece $x_i$, which preserves every candidate the pixels allow for the context layers.

\noindent\textbf{\hilo: high-to-low self-labelling.} Synthetic fonts do not cover a game's own display fonts,
but unlabelled videos of the same game do: their large text (titles, menus, banners) is easy to read. To exploit this
without any hand labels, \hilo reads every
line of at least 10\,px in six unlabelled ALO7 recordings at native resolution and keeps it only when our reading and
PP-OCR's agree exactly (6,398 of 8,680 lines). Each kept line is shrunk to 12 random heights in $[4.5, 9]$\,px (area
resampling, a random sub-pixel phase and JPEG compression), cut as at test time, and labelled with the agreed text. This yields 393k low-resolution
crops in the game's fonts. We fine-tune \clsnet with half of every batch drawn from them, and \cutnet on strips cut the same way
from lines of at least 12\,px, with the native-resolution cuts as targets. The recordings have no overlap with the benchmark videos.

\subsection{Context layer 1: coherence within a sentence}
\label{sec:layer1}

\noindent\textbf{Lattice product decoding.} The classifier estimates $q_i(c) \propto p(x_i \mid c)\, \pi(c)$, where $\pi$ is
the class prior of its training data. Dividing by $\pi$ therefore leaves a likelihood, while a language model supplies
the prior over strings. Following the noisy-channel formulation, \lpd reads a line as
\begin{equation}
  \begin{aligned}
  \hat c_{1:n} = \arg\max_{c_{1:n}} \; & \log P_{\mathrm{LM}}(c_{1:n} \mid \text{prompt}) \\
  & + \lambda \textstyle\sum_{i=1}^{n} \big[ \log q_i(c_i) - \beta \log \pi(c_i) \big],
  \end{aligned}
  \label{eq:lpd}
\end{equation}
where each $c_i$ ranges over the top $k=20$ candidates of piece $i$. With $\lambda=\beta=1$, \cref{eq:lpd} is
exactly the Bayes posterior $p(c_{1:n} \mid x_{1:n}) \propto P_{\mathrm{LM}}(c_{1:n}) \prod_i p(x_i \mid c_i)$ of a
noisy channel in which each piece is rendered from its character independently: the classifier, divided by its
training prior, is the channel, and the LLM is the source. The weights are therefore not a fusion heuristic tuned per
dataset but the values the derivation prescribes, and they are also the best on our dev split. This is the classical
decoder of speech and handwriting recognition~\cite{bourlard1994hybrid,wang2012hctr}, here with a modern LLM as the
prior. The LLM is Qwen3.5-9B-Base, frozen; its
vocabulary has a single-character token for 3,542 of our 3,548 classes. A beam search (width 8) runs over its tokens,
synchronised on character position. A token that spells several characters (a two-character word, say) is allowed
only if every character is among the candidates of the corresponding pieces, and hypotheses that spell the same string
through different tokenizations are merged by log-sum-exp. Symbols and junk keep the classifier's top-1.

\noindent\textbf{The document prompt.} To inform the language model of the game's names and vocabulary, we condition it on a
prompt consisting of a fixed header, \gl{the following is a piece of text from a game screen}, followed by up to 400
characters of the same video's native-resolution text, read by our own pipeline.

\noindent\textbf{Decode, do not correct.} Existing LLM-assisted OCR pipelines hand the language model the
recognizer's top-1 string, with or without the image, and ask it to fix it. This interface discards exactly what
matters at 5--7\,px: most errors there are confusions between characters that look alike only when blurred
(\cref{sec:matching}), which the top-1 string does not reveal and which neither a phonetic nor a generic glyph-confusion
model knows. The product keeps every candidate with its probability, which gives it two properties that correction
lacks. First, the two sources fail in different ways and each corrects the other's errors: the language model separates
look-alikes that the pixels cannot (piece 8 in \cref{fig:method}), and the pixels prevent the fluent words a language
model invents, since a reading outside the lattice has probability zero. Second, for the same reason, context can
inform the reading but cannot replace it: \lpd cannot copy a context line into the answer, which is what VLMs and
correctors do when given the same document prompt (\cref{sec:exp-main}). In practice, the product is the largest single
gain in \model (+8.8 points, \cref{tab:ladder}) and needs no game data, whereas correcting PP-OCRv6's string with the
same LLM leaves it where it was (53.5\% against 54.1\%) and, given the document prompt, lowers it to 35.6\%.

\subsection{Context layer 2: coherence within a video}
\label{sec:layer2}

A video uses only a few fonts, and the same character looks the same every time it appears. Inspired by how a human
reader recalls a word seen large in a title, \gbank exploits this regularity in two ways. First, the video's high-resolution lines, read almost perfectly by the same pipeline, are cut into characters,
downsampled to the target height (area resampling), and stored as labelled templates: \gbank, a glyph
library of \gl{this character, in this font, at this size}. Second, small pieces of the same video that look alike are
probably the same character, even when neither can be read alone.

For each target piece, \gbank retrieves its 24 nearest neighbours of the same height among the library templates and
the other small pieces of the video, by the similarity of their classifier distributions. A learned verifier estimates the
probability that each neighbour shows the same character. A small learned update (51k parameters) pools the
neighbours' evidence, weighted by the verifier, into a correction $\Delta_i(c)$ of the piece's log-distribution,
$\log q'_i(c) = \log q_i(c) + \Delta_i(c)$, over a candidate set that also includes labels proposed by confident
neighbours. The new lattice is decoded with \lpd.

We run two rounds. After the first decode, with the fixed prompt, every small piece has a reading in its own sentence; the second round then
also tells the update what each neighbour was read as, so that a character read correctly in one sentence helps read the
same character in another; the final decode adds the document prompt. The verifier and the update are trained on synthetic documents only, so \gbank never sees
benchmark text during training.

\subsection{Implementation}
\label{sec:impl}

Every component except the frozen LLM is small: \cutnet has 1.85M parameters (and its box refiner 1.28M), \clsnet
13M, and the \gbank verifier and update fewer than 1M together. All are trained on a single A40 GPU. The weights $\lambda=\beta=1$, the values the derivation prescribes, are also the best on a dev split of
rendered game dialogue; the same settings are used for every benchmark. Reading one line takes 0.69\,s on one A40, an upper
bound because each run decodes two prompt variants. Training data, architectures and schedules are in
\cref{app:impl}.

\section{Experiments}
\label{sec:exp}

\begin{table*}[t]
\centering
\small
\caption{\textbf{Main results on \bench.} Chinese accuracy rate (AR, \%) on the eight sub-benches at 5--7\,px (heights pooled; 6,126 lines, 52,236 Chinese characters). DP/ZB/AG/WS: the ALO7 games Dino Park, Zodiac Buddies, ALO7 General and Word Star; GL/GS/SL/SS: \textit{Genshin Impact} / \textit{Honkai: Star Rail}, large and small. Row groups: \model, \model without video context (no document prompt and no \gbank; \lpd with the fixed header only) and its classifier alone; baselines fine-tuned on our data; four LLM-assisted OCR pipelines; the strongest off-the-shelf systems, each with its best input. (SFT): fine-tuned on our training data; LLM: Qwen3.5-9B corrects the recognizer's string; VLM: Qwen3.5-9B corrects it while seeing the line image; Context: the document prompt \model gets. LA: exact line accuracy. Last column: \model minus the system, all sub-benches pooled, with a paired bootstrap 95\% interval over frames. All 19 off-the-shelf systems and all ten LLM-assisted pipelines are in \cref{tab:main_full,tab:llm}. Bold: best; underlined: second best.}
\label{tab:main}
\fitwidth{\begin{tabular}{lrrrrrrrrrrl}
\toprule
System & DP & ZB & AG & WS & GL & GS & SL & SS & All & LA & \model $-$ it \\
\midrule
\model & \textbf{64.4} & \textbf{71.3} & \textbf{61.5} & \textbf{68.6} & \textbf{91.9} & \textbf{89.4} & \textbf{94.2} & \textbf{94.7} & \textbf{84.2} & \textbf{51.8} & -- \\
\quad w/o video context & \underline{56.7} & \underline{70.9} & 55.4 & \underline{68.2} & \underline{90.6} & \underline{88.3} & \underline{93.0} & \underline{93.8} & \underline{82.6} & \underline{49.4} & $+1.6$ \scriptsize[+1.3, +2.0] \\
\quad classifier only & 48.4 & 66.3 & 44.7 & 56.2 & 82.7 & 80.5 & 83.1 & 83.3 & 73.8 & 33.4 & $+10.5$ \scriptsize[+9.9, +11.0] \\
\midrule
Qwen3.5-9B (SFT) & 53.0 & 68.3 & \underline{56.1} & 59.9 & 78.7 & 77.5 & 79.4 & 81.4 & 72.9 & 44.3 & $+11.3$ \scriptsize[+10.5, +12.1] \\
PP-OCRv6 (SFT) & 32.4 & 43.8 & 28.8 & 44.4 & 76.0 & 73.7 & 73.6 & 75.3 & 62.5 & 14.0 & $+21.7$ \scriptsize[+20.8, +22.6] \\
\midrule
Qwen3.5-9B (SFT) + LLM + Context & 46.6 & 17.5 & 51.5 & 57.2 & 75.1 & 74.7 & 74.5 & 75.7 & 62.4 & 35.5 & $+21.8$ \scriptsize[+20.1, +23.8] \\
PP-OCRv6 + VLM & 32.0 & 39.8 & 28.3 & 55.3 & 75.4 & 72.9 & 78.8 & 81.0 & 64.9 & 23.2 & $+19.3$ \scriptsize[+18.1, +20.5] \\
PP-OCRv6 + VLM + Context & 31.9 & 16.8 & 28.7 & 50.1 & 76.2 & 72.8 & 78.2 & 82.2 & 61.4 & 24.9 & $+22.8$ \scriptsize[+21.0, +24.7] \\
PP-OCRv6 (SFT) + VLM + Context & 39.6 & 44.4 & 34.5 & 59.8 & 81.2 & 79.8 & 82.5 & 85.9 & 70.0 & 29.3 & $+14.2$ \scriptsize[+13.1, +15.2] \\
\midrule
Qwen3.5-9B~\cite{qwen2026qwen35} & 32.6 & 36.8 & 29.0 & 56.3 & 76.8 & 75.2 & 80.9 & 82.6 & 65.9 & 23.7 & $+18.4$ \scriptsize[+17.0, +19.7] \\
Qwen3-VL-8B~\cite{bai2025qwen3vl} & 29.0 & 31.3 & 25.6 & 56.3 & 74.0 & 73.3 & 77.5 & 80.5 & 63.0 & 21.7 & $+21.3$ \scriptsize[+19.8, +22.8] \\
Qwen3-VL-32B~\cite{bai2025qwen3vl} & 13.4 & 27.9 & 19.9 & 50.9 & 69.0 & 68.3 & 72.0 & 76.6 & 57.2 & 18.5 & $+27.0$ \scriptsize[+25.3, +28.9] \\
HunyuanOCR (1B)~\cite{hunyuan2025ocr} & 19.8 & 25.5 & 18.7 & 42.2 & 67.0 & 64.9 & 68.3 & 71.8 & 54.3 & 11.6 & $+29.9$ \scriptsize[+28.4, +31.3] \\
PP-OCRv6 medium~\cite{zhang2026ppocrv6} & 20.1 & 34.1 & 18.5 & 44.3 & 64.4 & 63.8 & 66.1 & 68.7 & 54.1 & 10.7 & $+30.1$ \scriptsize[+28.9, +31.3] \\
PaddleOCR-VL-1.6 (0.9B)~\cite{cui2025paddleocrvl} & 25.5 & 30.2 & 19.0 & 44.3 & 63.8 & 60.0 & 65.5 & 71.4 & 53.8 & 11.5 & $+30.4$ \scriptsize[+29.2, +31.6] \\
MiniCPM-OCR 1.0~\cite{openbmb2026minicpmocr} & 27.5 & 31.1 & 23.5 & 42.6 & 62.4 & 60.8 & 63.7 & 66.4 & 52.8 & 12.0 & $+31.4$ \scriptsize[+30.4, +32.5] \\
GLM-OCR (0.9B)~\cite{duan2026glmocr} & 18.0 & 27.2 & 16.8 & 38.6 & 63.7 & 61.7 & 65.8 & 69.3 & 52.1 & 10.5 & $+32.2$ \scriptsize[+30.8, +33.6] \\
\bottomrule
\end{tabular}}
\end{table*}

\subsection{Setup}
\label{sec:setup}

\noindent\textbf{Baselines.} We compare with 19 off-the-shelf systems: expert recognizers (PP-OCRv5~\cite{cui2025paddleocr3} and
v6~\cite{zhang2026ppocrv6}, CnOCR~\cite{breezedeus2023cnocr}, EasyOCR~\cite{jaidedai2020easyocr},
Tesseract~\cite{smith2007tesseract}), OCR-specialised VLMs (PaddleOCR-VL~\cite{cui2025paddleocrvl},
HunyuanOCR~\cite{hunyuan2025ocr}, GLM-OCR~\cite{duan2026glmocr}, MiniCPM-OCR~\cite{openbmb2026minicpmocr},
GOT-OCR~2.0~\cite{wei2024got}, DeepSeek-OCR-2~\cite{wei2026deepseekocr2}) and general VLMs
(Qwen3.5-9B~\cite{qwen2026qwen35}, Qwen3-VL-8B and 32B~\cite{bai2025qwen3vl}, InternVL3.5~\cite{wang2025internvl35},
MiniCPM-V 4.5~\cite{yu2025minicpmv45}, LLaVA-OneVision-1.5~\cite{an2025llavaov15}). To separate
the effect of our design from that of our data, we also train the strongest expert and the strongest VLM on exactly the
data behind \clsnet: PP-OCRv6 fine-tuned with its official recipe, and Qwen3.5-9B with LoRA~\cite{hu2022lora}; we mark
them (SFT). Unless noted, PP-OCRv6 is its medium model.
We further run ten LLM-assisted OCR pipelines that correct a recognizer's draft: an LLM rewriting the string
(CLOCR-C~\cite{bourne2024clocrc}), a VLM seeing the image and the draft~\cite{greif2025multimodal}, and noisy-channel
Chinese spelling correction (simple-csc~\cite{zhou2024simplecsc}), with or without the same document prompt
\model gets. A pipeline is named by its recognizer and corrector: + LLM (Qwen3.5-9B rewrites the string), + VLM
(Qwen3.5-9B also sees the line image) and + Context (the document prompt). Finally, we apply super-resolution and then OCR (\cref{sec:sr}).

\noindent\textbf{Protocol.} Every system reads the same line window. Each off-the-shelf system is shown with its best input
(the raw pixels, or upscaled to 32 or 64\,px per em), chosen on \bench itself, which favours the baselines. Decoding is
greedy, with each model's documented recognition prompt. \model reads every frame of a video with the other half of the
frames at native resolution as context; baselines read lines without context unless stated. Details are in
\cref{app:baselines}.

\subsection{Main results}
\label{sec:exp-main}

\begin{figure*}[t]
  \centering
  \includegraphics[width=\textwidth]{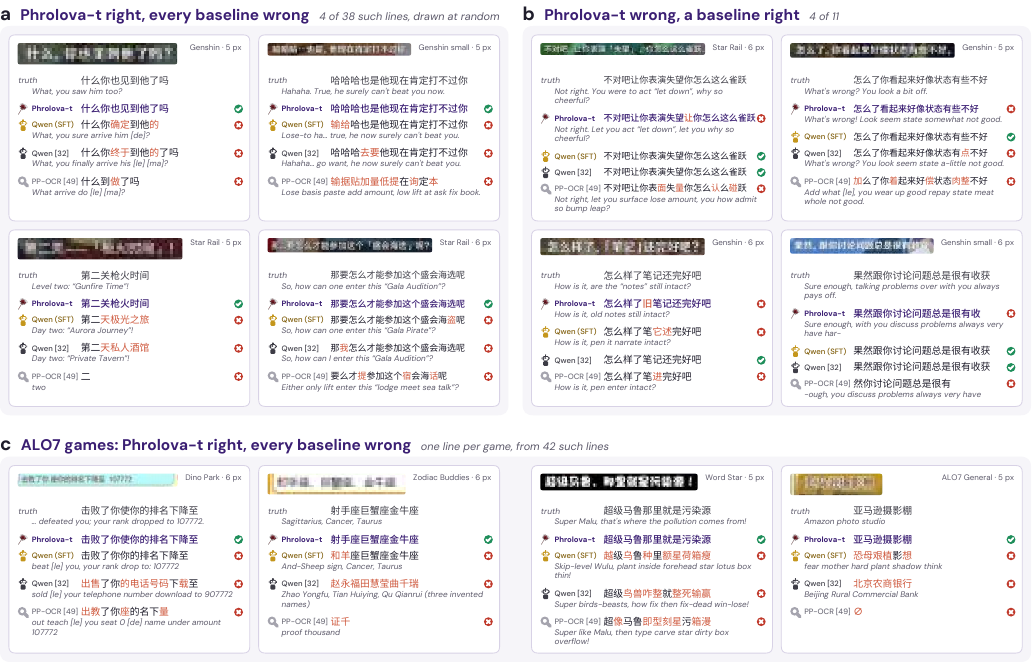}
  \caption{\textbf{Qualitative results.} (a, b) 5--6\,px Genshin and Star Rail lines, drawn at random from (a) the lines
  that \model reads exactly while Qwen3.5-9B (SFT), Qwen3.5-9B and PP-OCRv6 all miss them, and (b) the lines that \model misses
  while at least one of them reads them exactly. (c) One line from each ALO7 game, chosen among the lines of the first
  kind so that no player name is shown. Red: characters that differ from the truth; grey: literal English of each
  wrong reading. Game frames: research use only, no commercial use.}
  \label{fig:qual}
\end{figure*}

As shown in \cref{tab:main}, \model achieves the highest accuracy on every sub-bench,
reading 84.2\% of the Chinese characters and 51.8\% of the lines exactly. Specifically, it surpasses the strongest
baseline, Qwen3.5-9B with LoRA on our data (72.9\%), by \ci{+11.3}{+10.5}{+12.1} points, and the strongest
off-the-shelf system, Qwen3.5-9B (65.9\%), by 18.4 points.

\noindent\textbf{Design matters more than data.} Training on our data helps both baselines (+7.0 and +8.4 points): Qwen3.5-9B gains almost
only on the ALO7 games, whose fonts that data covers, and PP-OCRv6 on every sub-bench but Word Star. However, with the same data and without any language model,
the classifier of \model alone reads 73.8\% against 62.5\% for the fine-tuned PP-OCRv6, and \model without video context
reads 82.6\% against 72.9\% for Qwen3.5-9B (SFT). We therefore attribute most of the gain to the design rather than to the
data.

\noindent\textbf{Correcting a string does not recover what the recognizer lost.} The best LLM-assisted pipeline (PP-OCRv6
trained on our data, corrected by Qwen3.5-9B seeing the image, the draft and the document prompt) reads 70.0\%,
14.2 points below \model. Correcting the text alone leaves PP-OCRv6's draft where it was (53.5\% against 54.1\%), and so
does simple-csc, whose glyph-confusion model does not know which characters look alike at 5\,px (54.2\%). Given the
document prompt in their input, the VLMs and correctors copy context lines into their answer: the correction of
PP-OCRv6 falls from 53.5\% to 35.6\% without the image and from 64.9\% to 61.4\% with it (\cref{tab:llm}), and
Qwen3.5-9B, reading lines upscaled to 32\,px per em, from 65.3\% to 55.6\%.

\noindent\textbf{The dev bench and memorization.} Without Zodiac Buddies, the dev bench, \model leads the best baseline by
\ci{+12.3}{+11.2}{+13.2} points (\cref{tab:nodev}). To separate reading from remembering, we split the ALO7 lines by
whether their exact text occurs in the training recordings. \model leads Qwen3.5-9B (SFT) by \ci{+1.8}{+0.2}{+3.4} on the 1,715
seen lines, by \ci{+9.0}{+7.4}{+10.6} on the 1,608 unseen lines (the other 614 occur only inside longer training
lines), and by \ci{+13.8}{+12.7}{+14.8} on the miHoYo games,
for which neither saw any footage. These results demonstrate that the advantage grows where memorization cannot help.

\subsection{Where the advantage lives}
\label{sec:exp-where}

\begin{table}[t]
  \centering
  \small
  \setlength{\tabcolsep}{3.6pt}
  \fitwidth{\begin{tabular}{lrrrrrr}
    \toprule
    System & 5\,px & 6\,px & 7\,px & 8\,px & $\geq$9\,px & native \\
    \midrule
    \model & \textbf{77.8} & \textbf{89.1} & \textbf{91.7} & 93.3 & 95.3 & -- \\
    \quad w/o video context & \underline{75.3} & \underline{87.9} & \underline{91.2} & 93.4 & 95.2 & -- \\
    Qwen3.5-9B (SFT) & 52.7 & 80.3 & 90.2 & \textbf{94.6} & \textbf{97.9} & -- \\
    PP-OCRv6 (SFT) + VLM + Context & 52.8 & 80.4 & 88.7 & \underline{93.9} & \underline{97.7} & -- \\
    Qwen3.5-9B~\cite{qwen2026qwen35} & 48.3 & 76.7 & 86.3 & 93.1 & 97.2 & \underline{99.4} \\
    Qwen3-VL-8B~\cite{bai2025qwen3vl} & 44.6 & 72.9 & 84.7 & 91.0 & 96.3 & 98.6 \\
    HunyuanOCR~\cite{hunyuan2025ocr} & 34.9 & 63.3 & 79.5 & 88.3 & 95.6 & \textbf{99.5} \\
    PP-OCRv6 medium~\cite{zhang2026ppocrv6} & 31.6 & 62.9 & 79.2 & 88.3 & 95.1 & 99.2 \\
    \bottomrule
  \end{tabular}}
  \caption{\textbf{Chinese AR by line height} on the same 1,732 lines at every height ($\geq$9\,px pools 9, 10 and
  12\,px). Native: the system reading the native-resolution line.}
  \label{tab:height-bins}
\end{table}

\noindent\textbf{By height.} \cref{tab:height-bins} pools the sweep of \cref{fig:cliff} into bins. \model leads by
\ci{+25.1}{+23.1}{+27.3} points over Qwen3.5-9B (SFT) at 5\,px, by +8.8 at 6\,px and by +1.5 at 7\,px. On larger text, which is outside the scope of this paper, it trails by 1.3--2.6 points, as
expected: our reader commits to a closed codebook and a channel model built for tiny text, while a VLM's open
vocabulary and larger visual model win once the pixels suffice.

\noindent\textbf{By complexity.} The lead grows with the number of strokes: \model minus Qwen3.5-9B (SFT) is +6.7 points for
characters of 1--5 strokes, then +9.8, +13.1, +14.5, and +15.7 for characters of 15 strokes or more
(\cref{tab:complexity}), the characters that the ambiguity analysis found most often confused. On the established
benchmarks of \cref{sec:exp-outside} the same difference shows no such trend ($-1.6$ to $+0.9$ on BCTR and OmniDocBench, $-1.0$ to $+3.0$ on
Real-CE), so the trend reflects the game fonts and their context, not a general advantage on complex characters.

\subsection{Ablations}
\label{sec:exp-ablation}

\begin{table}[t]
  \centering
  \small
  \setlength{\tabcolsep}{4pt}
  \fitwidth{\begin{tabular}{lrrrr}
    \toprule
    Pipeline stage & ALO7 & miHoYo & All & LA \\
    \midrule
    \clsnet (synthetic data), top-1 & 41.3 & 82.9 & 68.2 & 22.9 \\
    \quad + \hilo self-labelling & 57.3 & 82.7 & 73.8 & 33.4 \\
    \quad + \lpd (layer 1) & 65.6 & 91.8 & 82.6 & 49.4 \\
    \quad + document prompt (layer 1) & \underline{67.1} & \underline{92.5} & \underline{83.5} & \underline{50.8} \\
    \quad + \gbank (layer 2) = \model & \textbf{68.1} & \textbf{93.0} & \textbf{84.2} & \textbf{51.8} \\
    \midrule
    \clsnet (synthetic) + \lpd + prompt & 55.4 & 92.8 & 79.6 & 40.7 \\
    \bottomrule
  \end{tabular}}
  \caption{\textbf{\model stage by stage} (Chinese AR, \%, 5--7\,px pooled). Each row adds one component to the
  row above. Last row: context layer~1 on the classifier trained on synthetic data only.}
  \label{tab:ladder}
\end{table}

\cref{tab:ladder} adds one component at a time.
\begin{itemize}[leftmargin=*,itemsep=1pt,topsep=2pt]
  \item \textbf{\hilo} adds +5.6 points overall and +16.0 on the ALO7 games, from unlabelled video alone. It changes
  nothing on the miHoYo games, whose footage it never saw: it adapts the reader to the game's fonts.
  \item \textbf{\lpd} is the largest step, +8.8 points, and it needs no game data: on the classifier
  trained only on synthetic text, layer~1 with the document prompt adds +11.4.
  \item \textbf{The document prompt} adds +0.9, and \textbf{\gbank} a further +0.7, with 95\% intervals
  above zero on both the ALO7 and the miHoYo games. The two context sources overlap; together they add +7.7 points on Dino Park and
  +6.1 on ALO7 General (\cref{tab:main}).
\end{itemize}

\subsection{Generalization to established benchmarks}
\label{sec:exp-outside}

\begin{table}[t]
  \centering
  \small
  \setlength{\tabcolsep}{3.4pt}
  \fitwidth{\begin{tabular}{lrrrr}
    \toprule
    & \multicolumn{2}{c}{BCTR + OmniDoc} & \multicolumn{2}{c}{Real-CE} \\
    \cmidrule(lr){2-3}\cmidrule(lr){4-5}
    System & 5--7\,px & 5\,px & 5\,px & 13\,mm \\
    \midrule
    \model & 79.6 & \textbf{70.7} & \underline{32.9} & 77.6 \\
    \quad w/o video context & 79.6 & \underline{70.2} & \textbf{35.9} & 78.2 \\
    Qwen3.5-9B (SFT) & 79.8 & 69.6 & 28.8 & 82.7 \\
    PP-OCRv6 (SFT) + VLM + Context & \textbf{80.5} & 68.6 & 32.7 & \underline{85.7} \\
    Qwen3.5-9B~\cite{qwen2026qwen35} & \underline{80.3} & 69.2 & 29.2 & \textbf{86.1} \\
    Qwen3-VL-8B~\cite{bai2025qwen3vl} & 77.9 & 64.8 & 23.7 & 84.6 \\
    PaddleOCR-VL~\cite{cui2025paddleocrvl} & 77.6 & 65.9 & 23.3 & 81.6 \\
    PP-OCRv6 medium~\cite{zhang2026ppocrv6} & 74.1 & 56.8 & 18.2 & 82.4 \\
    \bottomrule
  \end{tabular}}
  \caption{\textbf{Established benchmarks}, none of which \model was trained on (Chinese AR, \%). BCTR web, scene and
  document test lines and OmniDocBench pages shrunk to 5--7\,px with our protocol; Real-CE val, its 13\,mm shot shrunk
  to 5\,px and at its own resolution. Per-subset results are in \cref{tab:outside}.}
  \label{tab:outside-short}

\end{table}

\model and its components were trained on synthetic text and unlabelled ALO7 video only. We therefore test it, without
any adaptations, on established Chinese benchmarks: the web, scene and document test lines of BCTR~\cite{chen2021bctr}, the
simplified-Chinese pages of OmniDocBench~\cite{ouyang2025omnidocbench}, and the real low-resolution shots of
Real-CE~\cite{ma2023realce}. We shrink them to 5--7\,px with our protocol (\cref{app:outside}). This is a demanding
test of generalization: these benchmarks are public, and the VLMs we compare with were likely trained on them or on
their sources.

\cref{tab:outside-short} shows that \model is on par overall (79.6\% against 80.5\% for the best LLM-assisted pipeline
and 80.3\% for Qwen3.5-9B) and ahead at 5\,px, by +1.1 to +2.1 points over the three strongest systems. On real optical
blur, Real-CE shrunk to 5\,px, \model without video context reads 35.9\% against 29.2\% for Qwen3.5-9B
(\ci{+6.7}{+5.2}{+8.3}); pooled over 5--7\,px on Real-CE, however, the full \model trails the three best systems by 0.9--2.7
points. It loses where the pixels suffice or the domain is far from rendered text: on real scene
photos (BCTR scene, $-5.5$ against Qwen3.5-9B), at 7\,px ($-1.3$ to $-2.8$), and on Real-CE at its own resolution
(77.6\% against 86.1\%). When a document has no other text, as in BCTR and Real-CE, \gbank has nothing to match and
costs 1--3 points.

\subsection{Qualitative results}
\label{sec:exp-qual}

\cref{fig:qual} shows random lines from both sides; among the 5--6\,px miHoYo lines of 7--16 characters, 38 fall on
the first side and 11 on the second. Where the baselines fail, they often write fluent words that are not there, such
as \gl{Private Tavern} for \gl{Gunfire Time}, whose characters look alike at 5\,px (\cref{fig:qual}a): the language
prior of a VLM fills in what it cannot see. On the ALO7 games (\cref{fig:qual}c), Qwen3.5-9B turns the zodiac signs
\gl{Sagittarius, Cancer, Taurus} into three invented personal names and a studio sign into \gl{Beijing Rural
Commercial Bank}. In contrast, where \model fails, the error is usually a single character, replaced by a look-alike or
dropped at a weak cut. On the ALO7 games, the context sometimes misleads it: game vocabulary from the document prompt
occasionally replaces a correct reading (\cref{app:failures}).

\noindent\textbf{A complement to super-resolution.} SR and \model solve different problems. SR restores how text looks, and
at 5\,px it invents the characters (\cref{fig:sr}); \model restores which characters are there, without producing an
image. The two are complementary: \model's readings could serve as a text prior for SR, which we leave to future work.

\subsection{Cost and limitations}
\label{sec:exp-limits}

\model reads a line in 0.69\,s on one A40, against 0.12--0.14\,s for the 9B VLMs and 0.004\,s for PP-OCRv6
(\cref{tab:efficiency}); most of the time is the beam search of the frozen LLM. Its vocabulary is closed: it reads
none of the 0.6\% of benchmark characters outside its 3,500 Chinese characters. Finally, part of the ground truth was drafted by PP-OCR at native
resolution and then checked by hand.

\section{Conclusion}
\label{sec:conclusion}

In this work, we studied Chinese text at 5--7\,px, a regime that defeats today's OCR methods: the best VLM reads fewer
than half of the characters at 5\,px, and super-resolution invents strokes rather than recovering them. We argued that
the problem is not open-ended vision.
The pixels are ambiguous, but the answer is one of a few thousand characters seen through a known channel, so reading
becomes decoding against a finite codebook, with context to choose among the candidates the pixels allow. \model
implements this view with two layers of context, the sentence (\lpd) and the video (\gbank and \hilo), and on
\bench it reads 11 points more than the strongest baseline trained on the same data and 23 points more at 5\,px, while
staying on par on established benchmarks it never trained on or adapted to.

Beyond tiny Chinese, we believe the same reasoning applies wherever many classes, fine detail and tiny
renderings meet: other dense scripts, mathematical notation, small text in videos and, perhaps most valuably,
historical material. Blurred scans of old books and archives, worn steles and rubbings, and damaged inscriptions on
artefacts have lost their strokes, yet their text comes from a known script, often in a few woodblock or type styles,
next to legible passages in the same hand. Restoring such images with a generative model would invent strokes, as
super-resolution does at 5\,px; decoding them would keep only the readings the surviving marks allow and let a
language model of the period, and glyphs cut from the legible pages, choose among them, much as epigraphers do by
hand and as recent work does for ancient Greek inscriptions~\cite{assael2022ithaca}. For the repair of documents and
objects, the order would be reversed: the decoded text, with its candidates, would guide the restoration of the
missing strokes instead of following from it. This setting also sharpens the open questions of our work: rare and variant characters beyond the common codebook, degradations without a simple known channel, and classical rather
than modern language would become valuable future work.

{
    \small
    \bibliographystyle{ieeenat_fullname}
    \bibliography{main}
}

\clearpage
\appendix
\maketitlesupplementary

\noindent The appendix gives the benchmark's construction (\cref{app:bench}), implementation details
(\cref{app:impl}), the baseline protocol (\cref{app:baselines}), the protocol for the established benchmarks
(\cref{app:outside}), full results (\cref{app:more}), whole ALO7 frames (\cref{app:frames}) and failure cases
(\cref{app:failures}).

\section{The \bench benchmark}
\label{app:bench}

\noindent\textbf{Sources and statistics.} \cref{tab:benchstats} lists the eight sub-benches. The ALO7 sub-benches come from
screen recordings of four children's learning games: Dino Park and Zodiac Buddies (adventure games set in the same
virtual world), ALO7 General (the platform's shared screens) and Word Star (an English-word adventure, in which 61\%
of the characters are Latin letters or digits). The miHoYo sub-benches contain the voiced dialogue block (speaker name
and subtitle) of 1080p story videos of \textit{Genshin Impact} 7.1 and \textit{Honkai: Star Rail} 2.2. ``Large''
sub-benches take 150 frames from three consecutive videos, ``small'' ones 50 frames from one video, so they differ in
how much context a frame has.

\begin{table}[h]
  \centering
  \small
  \setlength{\tabcolsep}{3pt}
  \fitwidth{\begin{tabular}{llrrrl}
    \toprule
    Sub-bench & Abbr. & Frames & Lines & Chinese ch. & Ground truth \\
    \midrule
    Dino Park & DP & 32 & 1,055 & 4,230 & hand \\
    Zodiac Buddies (dev) & ZB & 106 & 1,614 & 6,929 & hand \\
    ALO7 General & AG & 8 & 349 & 1,419 & hand \\
    Word Star & WS & 150 & 919 & 5,823 & draft + hand \\
    Genshin, large & GL & 150 & 802 & 11,120 & draft + hand \\
    Genshin, small & GS & 50 & 266 & 3,703 & draft + hand \\
    Star Rail, large & SL & 150 & 833 & 13,629 & draft + hand \\
    Star Rail, small & SS & 50 & 288 & 5,383 & draft + hand \\
    \midrule
    All & & 696 & 6,126 & 52,236 & \\
    \bottomrule
  \end{tabular}}
  \caption{\textbf{Sub-benches of \bench.} Lines and characters are target lines with at least one Chinese character,
  summed over the 5, 6 and 7\,px sets. Draft + hand: drafted by PP-OCR at native resolution, then checked by hand.}
  \label{tab:benchstats}
\end{table}

\noindent\textbf{Construction.} Each frame is shrunk with area resampling until the median height of its lines is 5, 6 or
7\,px, and then JPEG-compressed at quality 75; text that was larger in the original stays larger. Lines of at most
8.5\,px are targets; lines of at least 10\,px are context. Every sub-bench has equal 5, 6 and 7\,px sets. Six sub-benches are
cross-fitted in two folds: one half of a video's frames is shrunk while the other half is kept at native resolution as
context, then the roles swap, so every frame is a target exactly once and a reader never sees the native version of a
target frame. Dino Park and Zodiac Buddies use a single split: 15 of 32 and 53 of 106 frames are shrunk targets, and the
others serve only as context. The miHoYo dialogue boxes are tight to the ink, so they are grown to 1.15 character heights, the
convention of the ALO7 boards. Speaker names in Star Rail are about 1.4 times the subtitle size and exceed 8.5\,px at
the 7\,px setting, which is why those sets have fewer lines.

\noindent\textbf{Resolution sweep.} For \cref{sec:cliff}, all eight sub-benches are rebuilt at 8, 9, 10 and 12\,px from the
same frames. The 1,732 lines that are targets at every height from 5 to 12\,px form the sweep, so every height compares
the same text. \cref{tab:e1height} gives the full results.

\noindent\textbf{Real 360p twin.} To test the simulated degradation, the miHoYo episodes were also read from a real
640$\times$360 stream of the same videos, frame-aligned with the 1080p frames. The same 801 dialogue lines (11,502
Chinese characters) are read from the real stream, where they are 7--13\,px tall, and from our simulation of it (the
1080p frames shrunk by 1/3 with our protocol). Over 27 systems the two accuracies correlate at $r = 0.994$ (Spearman
$\rho = 0.95$), and the simulation is harder by a median of 0.85 points, mostly for weak recognizers
(\cref{tab:realsim}). The twin shows that the simulation errs, if at all, against the baselines.

\begin{table*}[t]
\centering
\small
\caption{The same 801 dialogue lines read from the real 640$\times$360 stream and from our simulation of it (1080p $\times$1/3 + JPEG q75). The eight strongest systems and two \model variants shown; over 27 systems Pearson $r = 0.994$.}
\label{tab:realsim}
\fitwidth{\begin{tabular}{lrrr}
\toprule
System & real 360p & simulated & sim $-$ real \\
\midrule
Qwen3.5-9B~\cite{qwen2026qwen35} & 99.5 & 99.3 & $-0.2$  \\
Qwen3.5-9B (SFT) & 99.4 & 99.2 & $-0.3$ \\
Qwen3-VL-8B~\cite{bai2025qwen3vl} & 99.3 & 99.1 & $-0.3$  \\
Qwen3-VL-32B~\cite{bai2025qwen3vl} & 99.1 & 98.7 & $-0.4$  \\
HunyuanOCR (1B)~\cite{hunyuan2025ocr} & 99.1 & 98.5 & $-0.6$  \\
PP-OCRv6 medium~\cite{zhang2026ppocrv6} & 98.9 & 98.1 & $-0.8$  \\
PaddleOCR-VL-1.6 (0.9B)~\cite{cui2025paddleocrvl} & 98.9 & 97.9 & $-0.9$  \\
PP-OCRv5 server~\cite{cui2025paddleocr3} & 98.7 & 97.9 & $-0.9$  \\
\model w/o \gbank & 98.6 & 98.5 & $-0.1$ \\
\model w/o video context & 98.6 & 98.4 & $-0.1$ \\
\bottomrule
\end{tabular}}
\end{table*}

\noindent\textbf{Naturally low-resolution recordings.}
\label{app:blurry}
We also collected ALO7 recordings that are low-resolution at the source and are not benchmark videos: two phone recordings of
Zodiac Buddies, whose picture band is $360{\times}214$ pixels with text mostly 5--8\,px tall, and one $616{\times}360$
recording of the platform's screens. From 30 frames, PP-OCRv5 detected 615 lines on a 3$\times$ bicubic upscale. They
have no ground truth, so we check our readings by hand and use them qualitatively. Frames of the
first phone recording also served as backgrounds behind synthetic training text, but none of its text was used as a
label. We passed the frames through eight super-resolution methods: bicubic, Real-ESRGAN~\cite{wang2021realesrgan}
(general and anime), SwinIR~\cite{liang2021swinir}, the one-step diffusion model ODTSR, and three text SR models trained
on Chinese, MARCONet~\cite{li2023marconet}, MARCONet++ and DiffTSR~\cite{zhang2024difftsr}. The GAN models turn
5--8\,px strokes into blocky pseudo-strokes, and the diffusion models write clean, font-like characters that are often
confidently wrong (\gl{name} becomes \gl{more than}). PP-OCRv6 finds Chinese on 63\% of the lines in
the raw pixels but on only 27--29\% after GAN-type SR.

\noindent\textbf{Release and licensing.} The game frames remain the property of their publishers. We use them for research
only and show them in figures for non-commercial, scientific illustration. Code and model checkpoints are released;
the benchmark will be released upon copyright authorization from the publishers.

\noindent\textbf{Metrics.} Ground truth and predictions are normalised (NFKC, lower case, no whitespace), and every character
outside U+4E00--U+9FFF is removed from both before alignment; lines without a Chinese character are dropped (6,126 of
7,939 target lines remain). Chinese AR is $(N - S - D - I)/N$ over $N$ ground-truth characters with $S$
substitutions, $D$ deletions and $I$ insertions, so a model that adds text is penalised. LA is the share of lines whose
Chinese sequence is read exactly. Sub-benches pool their three heights. Intervals are 95\% paired bootstrap intervals
with frames as the resampling unit (lines for BCTR, pages for OmniDocBench), 2,000 resamples.

\section{Implementation details}
\label{app:impl}

\noindent\textbf{Synthetic training text.} The generator renders text blocks of 1--6 lines of 3--18 characters, with
punctuation, game-style tokens (\texttt{LV12}, \texttt{3/5}, \texttt{HP}) and, in 10\% of blocks, real sentences. Characters
are drawn with probability proportional to $f(c)^{0.3}$, a flattened version of their corpus frequency $f$, so that
every class is seen often. Text is rendered in 31 open-licence fonts (sans, serif, kai, rounded, display, handwriting and
8 pixel fonts) on backgrounds that are public store-page screenshots of 156 games without adult content (35\%), frames
of six ALO7 videos (30\%; they include the video of the Word Star sub-bench, internally named Word Star 1-3, and one of
the blurry phone recordings; they serve only as visual material, with the synthetic text placed away from their own
text and no text or label taken from them, so no benchmark text enters training) or procedural
boards (35\%), with outlines, shadows and translucent panels. Each block is degraded
three times, to 5--8, 9--13 and 14--20\,px, by a random sub-pixel shift, Gaussian blur (50\%), one of four resampling
kernels, 4:2:0 chroma subsampling (40\%), JPEG with quality 35--95 (50\%) and sensor noise (30\%). Character boxes
remain exact after degradation.

\noindent\textbf{\cutnet.} The line box is resampled to a strip 1.6\,em tall at 15\,px per em. A 2-D convolutional trunk
(widths 64--192) pools the height to one row, five dilated 1-D residual blocks (receptive field about $\pm$2\,em)
follow, and a $1{\times}1$ convolution predicts per-column cut and centre logits (1.85M parameters, binary
cross-entropy, 10 epochs on 778k synthetic strips with jittered boxes). Cuts are the peaks of the cut probability with
a minimum spacing of 0.3\,em; pieces without centre evidence are dropped. A second network of the same design (widths
48--160, 1.28M parameters) with a head that regresses the box error, trained the same way with stronger box jitter,
first corrects the line box. \hilo fine-tunes the cutter for 4 epochs on 102k self-labelled strips mixed with a quarter
of the second synthetic strip set.

\noindent\textbf{\clsnet.} A CIFAR-style ResNet-18 (widths 64--512, 13.0M parameters) reads the central 1.3\,em of a
$32{\times}32$ crop of 1.6\,em. It is trained with cross-entropy and no label smoothing, to keep its probabilities
calibrated, on 18.3M synthetic crops (571k steps of 512, AdamW, learning rate $2{\times}10^{-3}$, cosine schedule),
with simulated segmentation error, scale, colour and noise augmentation. Junk crops (halves of two neighbouring
characters, or background past the line's end) teach the \texttt{<junk>} class. \hilo then fine-tunes it for 20k steps
(learning rate $3{\times}10^{-4}$) with half of every batch from the 393k self-labelled game crops.

\noindent\textbf{\hilo data.} Lines of at least 10\,px (12\,px for the cut strips) in six unlabelled ALO7 recordings are
read at native resolution by our synthetic-only pipeline and by PP-OCR and kept only when the two readings agree exactly (6,398 of 8,680 lines, 2,838 distinct
texts, 1,291 distinct characters). Each kept line is shrunk 12 times to a random height in $[4.5, 9]$\,px (area
resampling, random sub-pixel phase, JPEG quality 55--90), its box is refined and it is cut as at test time; a piece
is labelled with the agreed character when it lies within 0.3\,em of the mapped native cut (87\% of copies), and
otherwise the mapped native cuts are used. The recordings are disjoint from the benchmark videos, but they show the
same virtual world as Dino Park and Zodiac Buddies, which is why we report seen and unseen lines separately
(\cref{sec:exp-main}).

\noindent\textbf{\lpd.} The language model is Qwen3.5-9B-Base in bf16, frozen. The beam (width 8) is synchronised on
character position; a token extends a hypothesis when each of its characters is among the corresponding pieces' top
20, and hypotheses spelling the same string through different tokenizations are merged by log-sum-exp, which
approximates the string probability. Positions with no feasible token keep the classifier's top-1. The class prior
$\pi$ counts the classifier's training labels with add-one smoothing. Each line is decoded on its own, and every beam
hypothesis carries the model state after its tokens, so extending it costs one token. The first round uses the fixed
prompt; the second adds the document prompt, up to 400 characters of the video's lines of at least 10\,px in document
order, read by our classifier.

\noindent\textbf{\gbank.} For every piece we keep the classifier's 512-d feature and its top-20 candidates. Library templates
are the pieces of the video's lines of at least 10\,px, downsampled to every target height from 5 to 8\,px below their
own with area resampling. Each target piece retrieves its 24 nearest neighbours of the same height among the library
templates and the video's other small pieces, by the Bhattacharyya coefficient of the output distributions. A
verifier (0.38M parameters) predicts whether two pieces show the same character from their crops, features and the
distribution and feature similarities; it is trained on 88M pairs from 2,400 synthetic single-font documents, mined as
each target's top 24 neighbours by distribution similarity plus its top 8 by feature cosine. The update model (51k parameters per round) encodes, for every
neighbour and candidate label, the neighbour's log-probability, whether the label is its top-1, the verifier logit,
the two similarities and whether it is a library template; it pools these codes over neighbours (mean, sum, max) and
outputs a correction $\Delta_i(c)$ over 32 candidates. Each candidate's punctuation flag is also an input, because punctuation marks are near-identical blobs at this size.
In round~2 it also sees whether each neighbour's round-1 reading equals $c$. The round-1 update is trained on the 2,400
synthetic documents; the round-2 update starts from it and is trained on 1,200 of them, decoded with the language model
as at test time. The verifier and both updates were trained on the features of the synthetic-only \clsnet (before
\hilo) and are applied unchanged to the final one.

\noindent\textbf{Compute.} Every model is trained on a single NVIDIA A40 GPU; \cutnet takes under an hour and the \hilo
fine-tuning of \clsnet about 20 minutes. Running all systems on the benchmark took about four hours on 11 A40s.

\section{Baselines}
\label{app:baselines}

\noindent\textbf{Inputs.} Every system reads the same window: the line box plus 0.5\,em left and right and 0.25\,em above
and below. ``raw'' is that window at its low resolution; ``up32'' and ``up64'' resize it bicubically so that one em is
32 or 64\,px, because a 5\,px line is smaller than one vision-transformer patch. Expert recognizers are run raw and
up32, VLMs up32 and up64; each system is reported with its best input. Each system also reads the native-resolution
window as a reference.

\noindent\textbf{Prompts and decoding.} OCR-specialised models use their documented recognition prompt. General VLMs use
\gl{Read the text in this image exactly as written (it is usually Chinese). Output only the text, with no explanation,
translation or formatting.} Decoding is greedy, thinking is off, the output budget is two tokens per em of box width,
and every output receives the same clean-up.

\noindent\textbf{Baselines trained on our data.} Both use exactly the data behind \clsnet: every line of the synthetic sets
(1.56M line windows at all heights) plus the 6,398 self-labelled game lines shrunk to 12 random heights each (76,716 usable windows), half
of each batch from each source, cut like the baselines' benchmark input. PP-OCRv6 medium is fine-tuned from its
official weights with the official recipe for 8 epochs (learning rate $2{\times}10^{-4}$). Qwen3.5-9B gets LoRA (rank
16 on every linear layer of the vision tower and the language model) with the general prompt and up32 input, for
2,000 steps of 64 lines (learning rate $10^{-4}$).

\noindent\textbf{LLM-assisted OCR.} We follow the published recipes on the strongest drafts, with greedy decoding and the
same output budget. Text correction~\cite{bourne2024clocrc,kanerva2025nofreelunch}: an instruct LLM rewrites the draft.
VLM correction~\cite{greif2025multimodal}: the VLM sees the line image (up64) and the draft. simple-csc~\cite{zhou2024simplecsc}:
training-free noisy-channel correction with a base LLM (Qwen2.5-7B, as recommended) and a phonetic and glyph-shape
distortion model, with insertions and deletions enabled. ``+ context'' adds the document prompt \model gets.
\cref{tab:llm} gives all ten pipelines.

\begin{table*}[t]
\centering
\small
\caption{LLM-assisted OCR baselines on the game sub-benches (Chinese AR, \%). Ordered by name. (SFT): fine-tuned on our training data; LLM: Qwen3.5-9B corrects the recognizer's string; VLM: Qwen3.5-9B corrects it while seeing the line image; Context: the document prompt \model gets. Qwen3-VL-32B: that VLM as the corrector; simple-csc: noisy-channel Chinese spelling correction with a base LLM~\cite{zhou2024simplecsc}.}
\label{tab:llm}
\fitwidth{\begin{tabular}{lrrrrl}
\toprule
LLM-assisted OCR & ALO7 & miHoYo & All & LA & \model $-$ it \\
\midrule
Qwen3.5-9B (SFT) + LLM + Context & 39.4 & 74.9 & 62.4 & \underline{35.5} & $+21.8$ \scriptsize[+20.1, +23.8] \\
PP-OCRv6 + LLM & 31.9 & 65.3 & 53.5 & 11.7 & $+30.7$ \scriptsize[+29.5, +31.9] \\
PP-OCRv6 + LLM + Context & -5.9 & 58.2 & 35.6 & 12.4 & $+48.6$ \scriptsize[+45.2, +52.6] \\
PP-OCRv6 + VLM & 42.0 & 77.4 & 64.9 & 23.2 & $+19.3$ \scriptsize[+18.1, +20.5] \\
PP-OCRv6 + VLM + Context & 31.7 & 77.6 & 61.4 & 24.9 & $+22.8$ \scriptsize[+21.0, +24.7] \\
PP-OCRv6 + Qwen3-VL-32B + Context & 35.4 & 73.8 & 60.3 & 23.5 & $+24.0$ \scriptsize[+22.4, +25.7] \\
PP-OCRv6 + simple-csc & 32.6 & 66.0 & 54.2 & 11.4 & $+30.0$ \scriptsize[+28.8, +31.2] \\
PP-OCRv6 + simple-csc + Context & 33.0 & 66.3 & 54.6 & 11.7 & $+29.6$ \scriptsize[+28.4, +30.9] \\
PP-OCRv6 (SFT) + LLM + Context & 8.2 & 69.7 & 48.0 & 16.6 & $+36.2$ \scriptsize[+33.1, +39.7] \\
PP-OCRv6 (SFT) + VLM + Context & \underline{47.4} & \underline{82.3} & \underline{70.0} & 29.3 & $+14.2$ \scriptsize[+13.1, +15.2] \\
\midrule
\model & \textbf{68.1} & \textbf{93.0} & \textbf{84.2} & \textbf{51.8} & -- \\
\bottomrule
\end{tabular}}
\end{table*}

\section{Established benchmarks}
\label{app:outside}

\noindent\textbf{BCTR.} We sample 1,000 test lines per split (web, scene, document) with seed 0 among horizontal lines at
least 24\,px tall with 2--25 characters, at least two of them Chinese. The web and scene splits come from a
label-cleaned public copy of the test set; the document split is the original. Each line is its own document, so \model runs without context.

\noindent\textbf{OmniDocBench.} We take 70 simplified-Chinese pages, up to 8 per data source, excluding historical documents. The release has block transcripts
but no line boxes, so lines are detected at native resolution and each block's human transcript is split at the line
boundaries of its alignment with a native-resolution reading; blocks whose reading is more than 5\% CER from the
transcript are dropped (879 of 1,098 kept). The human transcript is always what is scored. Each page is shrunk so that
its median line is the target height; lines of at least 10\,px (mostly titles) are the page's context.

\noindent\textbf{Real-CE.} We use the 300 scenes of the validation split and its 2,267 horizontal lines with 2--25
characters (15,892 Chinese characters). The 13\,mm shot at its real resolution has lines of 5--30\,px (median
10--15\,px); we also shrink the same windows to 5--7\,px with our protocol, adding our pixel loss to real optical blur.
Labels and predictions are converted from traditional to simplified Chinese before scoring.

\begin{table*}[t]
\centering
\small
\caption{Established benchmarks (Chinese AR, \%). BCTR~\cite{chen2021bctr} web / scene / document test lines and OmniDocBench simplified-Chinese pages shrunk to 5--7\,px with our protocol; the 5\,px column pools all four. Real-CE~\cite{ma2023realce} val: the 13\,mm shot shrunk to 5--7\,px, and at its real resolution (13\,mm). No context exists on BCTR and Real-CE. (SFT), VLM and Context as in \cref{tab:main}.}
\label{tab:outside}
\fitwidth{\begin{tabular}{lrrrrrrrr}
\toprule
System & \multicolumn{5}{c}{BCTR and OmniDocBench, 5--7\,px} & \multicolumn{3}{c}{Real-CE} \\ \cmidrule(lr){2-6}\cmidrule(lr){7-9} & web & scene & doc. & OmniDoc & 5\,px & 5--7\,px & 5\,px & 13\,mm \\
\midrule
\model & 53.1 & 17.8 & \underline{40.9} & 93.2 & \textbf{70.7} & 52.2 & \underline{32.9} & 77.6 \\
\quad w/o video context & 54.2 & 19.2 & \textbf{43.2} & 92.8 & \underline{70.2} & \underline{54.2} & \textbf{35.9} & 78.2 \\
\midrule
Qwen3.5-9B (SFT) & 51.0 & 19.4 & 36.5 & \underline{94.2} & 69.6 & 51.4 & 28.8 & 82.7 \\
PP-OCRv6 + VLM + Context & \underline{56.3} & \textbf{27.9} & 39.0 & 93.3 & 67.9 & 53.6 & 29.9 & 85.6 \\
PP-OCRv6 (SFT) + VLM + Context & \textbf{56.4} & \underline{24.7} & 40.3 & 93.7 & 68.6 & \textbf{54.9} & 32.7 & \underline{85.7} \\
\midrule
Qwen3.5-9B~\cite{qwen2026qwen35} & 48.0 & 23.3 & 39.9 & \textbf{94.3} & 69.2 & 53.1 & 29.2 & \textbf{86.1} \\
Qwen3-VL-8B~\cite{bai2025qwen3vl} & 43.3 & 17.5 & 37.8 & 92.3 & 64.8 & 48.1 & 23.7 & 84.6 \\
PaddleOCR-VL-1.6 (0.9B)~\cite{cui2025paddleocrvl} & 42.4 & 16.5 & 34.3 & 92.6 & 65.9 & 44.8 & 23.3 & 81.6 \\
MiniCPM-OCR 1.0~\cite{openbmb2026minicpmocr} & 45.3 & 18.4 & 30.4 & 88.5 & 60.6 & 44.5 & 24.8 & 81.1 \\
GLM-OCR (0.9B)~\cite{duan2026glmocr} & 41.7 & 17.0 & 28.4 & 89.4 & 60.7 & 44.0 & 23.3 & 79.9 \\
HunyuanOCR (1B)~\cite{hunyuan2025ocr} & 35.5 & 14.9 & 23.5 & 85.0 & 52.8 & 38.9 & 15.3 & 81.9 \\
PP-OCRv6 medium~\cite{zhang2026ppocrv6} & 52.5 & 24.5 & 31.2 & 86.9 & 56.8 & 42.7 & 18.2 & 82.4 \\
PP-OCRv6 (SFT) & 48.8 & 17.1 & 29.2 & 84.4 & 55.5 & 40.5 & 19.9 & 78.9 \\
PP-OCRv5 server~\cite{cui2025paddleocr3} & 49.0 & 21.4 & 25.4 & 82.4 & 50.2 & 35.8 & 12.3 & 82.2 \\
\bottomrule
\end{tabular}}
\end{table*}

\section{Further results}
\label{app:more}

\begin{figure}[h]
  \centering
  \includegraphics[width=\columnwidth]{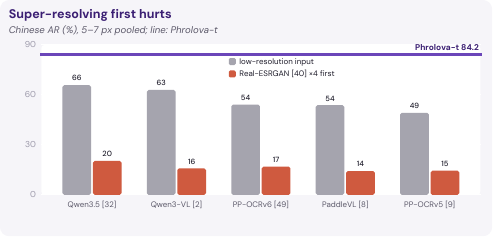}
  \caption{\textbf{Super-resolution then OCR.} Chinese AR of five readers from the low-resolution input and after
  Real-ESRGAN $\times$4, all eight sub-benches, 5--7\,px; the line marks \model.}
  \label{fig:srthenocr}
\end{figure}

\begin{figure}[h]
  \centering
  \includegraphics[width=\columnwidth]{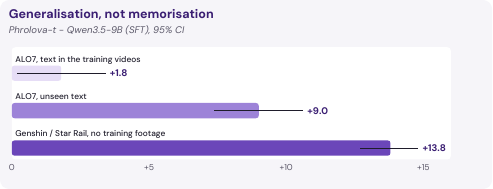}
  \caption{\textbf{Reading, not remembering.} \model minus Qwen3.5-9B (SFT), on ALO7 lines whose exact text
  occurs in the training recordings, on unseen ALO7 lines, and on the miHoYo games, with 95\% intervals.}
  \label{fig:seenunseen}
\end{figure}

\cref{tab:main_full} lists all 19 off-the-shelf systems. \cref{tab:height} splits the main result by height and game
family, \cref{tab:e1height} gives the full resolution sweep and \cref{tab:nodev} the main result without the dev bench.
\cref{tab:complexity} gives accuracy by stroke count, \cref{tab:ambiguity} the learning-free ambiguity analysis, and
\cref{tab:efficiency} the cost per line. \cref{fig:srthenocr} shows super-resolution then OCR for five readers, and
\cref{fig:seenunseen} the lead over Qwen3.5-9B (SFT) on seen and unseen text.

\begin{table*}[t]
\centering
\small
\caption{Chinese accuracy rate (AR, \%) on the eight game sub-benches at 5--7\,px (heights pooled; 6,126 lines, 52,236 Chinese characters). DP/ZB/AG/WS: the ALO7 games Dino Park, Zodiac Buddies, ALO7 General and Word Star; GL/GS/SL/SS: \textit{Genshin Impact} / \textit{Honkai: Star Rail}, large and small. Rows 2--3: \model without video context (no document prompt and no \gbank; \lpd with the fixed header only) and its classifier alone. LA: exact line accuracy. Last column: \model minus the system, all sub-benches pooled, with a paired bootstrap 95\% interval over frames. Off-the-shelf systems are shown with their best input (up32, up64 or raw). (SFT), LLM, VLM and Context as in \cref{tab:main}.}
\label{tab:main_full}
\fitwidth{\begin{tabular}{lrrrrrrrrrrl}
\toprule
System & DP & ZB & AG & WS & GL & GS & SL & SS & All & LA & \model $-$ it \\
\midrule
\model & \textbf{64.4} & \textbf{71.3} & \textbf{61.5} & \textbf{68.6} & \textbf{91.9} & \textbf{89.4} & \textbf{94.2} & \textbf{94.7} & \textbf{84.2} & \textbf{51.8} & -- \\
\quad w/o video context & \underline{56.7} & \underline{70.9} & 55.4 & \underline{68.2} & \underline{90.6} & \underline{88.3} & \underline{93.0} & \underline{93.8} & \underline{82.6} & \underline{49.4} & $+1.6$ \scriptsize[+1.3, +2.0] \\
\quad classifier only & 48.4 & 66.3 & 44.7 & 56.2 & 82.7 & 80.5 & 83.1 & 83.3 & 73.8 & 33.4 & $+10.5$ \scriptsize[+9.9, +11.0] \\
\midrule
Qwen3.5-9B (SFT) & 53.0 & 68.3 & \underline{56.1} & 59.9 & 78.7 & 77.5 & 79.4 & 81.4 & 72.9 & 44.3 & $+11.3$ \scriptsize[+10.5, +12.1] \\
PP-OCRv6 (SFT) & 32.4 & 43.8 & 28.8 & 44.4 & 76.0 & 73.7 & 73.6 & 75.3 & 62.5 & 14.0 & $+21.7$ \scriptsize[+20.8, +22.6] \\
\midrule
Qwen3.5-9B (SFT) + LLM + Context & 46.6 & 17.5 & 51.5 & 57.2 & 75.1 & 74.7 & 74.5 & 75.7 & 62.4 & 35.5 & $+21.8$ \scriptsize[+20.1, +23.8] \\
PP-OCRv6 + VLM & 32.0 & 39.8 & 28.3 & 55.3 & 75.4 & 72.9 & 78.8 & 81.0 & 64.9 & 23.2 & $+19.3$ \scriptsize[+18.1, +20.5] \\
PP-OCRv6 + VLM + Context & 31.9 & 16.8 & 28.7 & 50.1 & 76.2 & 72.8 & 78.2 & 82.2 & 61.4 & 24.9 & $+22.8$ \scriptsize[+21.0, +24.7] \\
PP-OCRv6 (SFT) + VLM + Context & 39.6 & 44.4 & 34.5 & 59.8 & 81.2 & 79.8 & 82.5 & 85.9 & 70.0 & 29.3 & $+14.2$ \scriptsize[+13.1, +15.2] \\
\midrule
Qwen3.5-9B~\cite{qwen2026qwen35} & 32.6 & 36.8 & 29.0 & 56.3 & 76.8 & 75.2 & 80.9 & 82.6 & 65.9 & 23.7 & $+18.4$ \scriptsize[+17.0, +19.7] \\
Qwen3-VL-8B~\cite{bai2025qwen3vl} & 29.0 & 31.3 & 25.6 & 56.3 & 74.0 & 73.3 & 77.5 & 80.5 & 63.0 & 21.7 & $+21.3$ \scriptsize[+19.8, +22.8] \\
Qwen3-VL-32B~\cite{bai2025qwen3vl} & 13.4 & 27.9 & 19.9 & 50.9 & 69.0 & 68.3 & 72.0 & 76.6 & 57.2 & 18.5 & $+27.0$ \scriptsize[+25.3, +28.9] \\
HunyuanOCR (1B)~\cite{hunyuan2025ocr} & 19.8 & 25.5 & 18.7 & 42.2 & 67.0 & 64.9 & 68.3 & 71.8 & 54.3 & 11.6 & $+29.9$ \scriptsize[+28.4, +31.3] \\
PP-OCRv6 medium~\cite{zhang2026ppocrv6} & 20.1 & 34.1 & 18.5 & 44.3 & 64.4 & 63.8 & 66.1 & 68.7 & 54.1 & 10.7 & $+30.1$ \scriptsize[+28.9, +31.3] \\
PaddleOCR-VL-1.6 (0.9B)~\cite{cui2025paddleocrvl} & 25.5 & 30.2 & 19.0 & 44.3 & 63.8 & 60.0 & 65.5 & 71.4 & 53.8 & 11.5 & $+30.4$ \scriptsize[+29.2, +31.6] \\
MiniCPM-OCR 1.0~\cite{openbmb2026minicpmocr} & 27.5 & 31.1 & 23.5 & 42.6 & 62.4 & 60.8 & 63.7 & 66.4 & 52.8 & 12.0 & $+31.4$ \scriptsize[+30.4, +32.5] \\
GLM-OCR (0.9B)~\cite{duan2026glmocr} & 18.0 & 27.2 & 16.8 & 38.6 & 63.7 & 61.7 & 65.8 & 69.3 & 52.1 & 10.5 & $+32.2$ \scriptsize[+30.8, +33.6] \\
PP-OCRv6 small~\cite{zhang2026ppocrv6} & 18.5 & 30.8 & 16.8 & 40.6 & 59.9 & 59.2 & 60.9 & 63.7 & 50.0 & 7.4 & $+34.3$ \scriptsize[+33.2, +35.3] \\
PP-OCRv5 server~\cite{cui2025paddleocr3} & 19.7 & 30.9 & 16.8 & 40.0 & 59.9 & 58.1 & 59.4 & 60.9 & 49.3 & 11.7 & $+35.0$ \scriptsize[+33.9, +36.0] \\
PP-OCRv5 mobile~\cite{cui2025paddleocr3} & 10.9 & 22.6 & 11.1 & 30.1 & 49.6 & 47.7 & 48.6 & 51.5 & 39.5 & 5.5 & $+44.8$ \scriptsize[+43.7, +45.8] \\
CnOCR 2.3~\cite{breezedeus2023cnocr} & 10.6 & 19.5 & 9.2 & 26.7 & 43.9 & 43.9 & 44.3 & 45.2 & 35.3 & 4.5 & $+48.9$ \scriptsize[+47.9, +49.9] \\
GOT-OCR 2.0 (0.6B)~\cite{wei2024got} & 5.8 & 16.0 & 7.6 & 27.8 & 41.6 & 43.1 & 39.5 & 40.1 & 32.3 & 7.4 & $+52.0$ \scriptsize[+50.7, +53.4] \\
LLaVA-OneVision-1.5-8B~\cite{an2025llavaov15} & 3.4 & 9.4 & 5.8 & 18.5 & 25.9 & 26.9 & 23.6 & 27.8 & 20.2 & 3.4 & $+64.1$ \scriptsize[+62.6, +65.6] \\
Tesseract 5~\cite{smith2007tesseract} & 1.5 & 2.6 & 1.1 & 12.1 & 21.0 & 18.5 & 18.4 & 19.9 & 14.5 & 0.7 & $+69.7$ \scriptsize[+68.4, +71.1] \\
DeepSeek-OCR-2 (3B)~\cite{wei2026deepseekocr2} & -8.0 & -4.7 & -5.5 & 7.6 & 24.6 & 22.1 & 19.1 & 20.3 & 13.3 & 0.4 & $+70.9$ \scriptsize[+69.0, +72.9] \\
MiniCPM-V 4.5 (8B)~\cite{yu2025minicpmv45} & 4.5 & 8.8 & 5.2 & 18.1 & 12.9 & 14.5 & 11.9 & 10.6 & 11.7 & 6.1 & $+72.6$ \scriptsize[+70.9, +74.3] \\
EasyOCR~\cite{jaidedai2020easyocr} & 0.3 & 0.7 & 0.1 & 3.1 & 5.5 & 5.5 & 4.7 & 4.3 & 3.7 & 0.2 & $+80.5$ \scriptsize[+78.8, +82.1] \\
InternVL3.5-8B~\cite{wang2025internvl35} & 0.8 & 0.1 & -1.8 & 1.4 & -2.1 & -2.5 & -0.3 & 2.1 & -0.3 & 1.5 & $+84.5$ \scriptsize[+82.6, +86.4] \\
\bottomrule
\end{tabular}}
\end{table*}

\begin{table*}[t]
\centering
\small
\caption{Chinese AR (\%) by line height on the game sub-benches. (SFT), VLM and Context as in \cref{tab:main}.}
\label{tab:height}
\fitwidth{\begin{tabular}{lrrrrrr}
\toprule
System & \multicolumn{3}{c}{ALO7} & \multicolumn{3}{c}{miHoYo} \\ \cmidrule(lr){2-4}\cmidrule(lr){5-7} & 5\,px & 6\,px & 7\,px & 5\,px & 6\,px & 7\,px \\
\midrule
\model & \textbf{56.0} & \textbf{68.8} & \textbf{76.4} & \textbf{85.6} & \textbf{95.8} & \textbf{97.7} \\
\quad w/o video context & \underline{53.9} & \underline{65.3} & \underline{74.7} & \underline{82.7} & \underline{95.4} & \underline{97.4} \\
Qwen3.5-9B (SFT) & 46.8 & 61.0 & 72.1 & 56.2 & 86.6 & 95.0 \\
PP-OCRv6 (SFT) + VLM + Context & 31.0 & 46.8 & 60.2 & 61.2 & 89.2 & 96.7 \\
Qwen3.5-9B~\cite{qwen2026qwen35} & 24.7 & 41.6 & 53.8 & 57.0 & 85.9 & 94.7 \\
Qwen3-VL-8B~\cite{bai2025qwen3vl} & 23.2 & 36.5 & 51.3 & 52.3 & 83.1 & 93.8 \\
PaddleOCR-VL-1.6 (0.9B)~\cite{cui2025paddleocrvl} & 18.3 & 29.8 & 46.5 & 38.0 & 70.7 & 87.4 \\
PP-OCRv6 (SFT) & 25.9 & 39.0 & 52.1 & 52.2 & 79.6 & 92.3 \\
PP-OCRv6 medium~\cite{zhang2026ppocrv6} & 17.4 & 33.9 & 43.6 & 37.1 & 71.2 & 88.9 \\
\bottomrule
\end{tabular}}
\end{table*}

\begin{table*}[t]
\centering
\small
\caption{Chinese AR (\%) by line height on the eight game sub-benches, the same 1,732 lines at every height ($\geq$\,9\,px pools 9, 10 and 12\,px). Native: the system's reading at the original resolution. (SFT), VLM and Context as in \cref{tab:main}.}
\label{tab:e1height}
\fitwidth{\begin{tabular}{lrrrrrr}
\toprule
System & 5\,px & 6\,px & 7\,px & 8\,px & $\geq$\,9\,px & native \\
\midrule
\model & \textbf{77.8} & \textbf{89.1} & \textbf{91.7} & 93.3 & 95.3 & -- \\
\quad w/o video context & \underline{75.3} & \underline{87.9} & \underline{91.2} & 93.4 & 95.2 & -- \\
Qwen3.5-9B (SFT) & 52.7 & 80.3 & 90.2 & \textbf{94.6} & \textbf{97.9} & -- \\
PP-OCRv6 (SFT) + VLM + Context & 52.8 & 80.4 & 88.7 & \underline{93.9} & \underline{97.7} & -- \\
PP-OCRv6 (SFT) & 45.2 & 70.4 & 82.7 & 89.0 & 93.8 & -- \\
Qwen3.5-9B~\cite{qwen2026qwen35} & 48.3 & 76.7 & 86.3 & 93.1 & 97.2 & \underline{99.4} \\
Qwen3-VL-8B~\cite{bai2025qwen3vl} & 44.6 & 72.9 & 84.7 & 91.0 & 96.3 & 98.6 \\
PaddleOCR-VL-1.6 (0.9B)~\cite{cui2025paddleocrvl} & 32.7 & 61.7 & 78.3 & 87.4 & 94.9 & 99.2 \\
HunyuanOCR (1B)~\cite{hunyuan2025ocr} & 34.9 & 63.3 & 79.5 & 88.3 & 95.6 & \textbf{99.5} \\
PP-OCRv6 medium~\cite{zhang2026ppocrv6} & 31.6 & 62.9 & 79.2 & 88.3 & 95.1 & 99.2 \\
PP-OCRv5 server~\cite{cui2025paddleocr3} & 23.7 & 56.5 & 76.7 & 87.5 & 95.4 & 98.3 \\
\bottomrule
\end{tabular}}
\end{table*}

\begin{table*}[t]
\centering
\small
\caption{Main results without the dev bench (Zodiac Buddies, \textit{ZB}, which informed development decisions): Chinese AR (\%) on the other seven game sub-benches, 5--7\,px (4,512 lines, 45,307 Chinese characters). ``All 8'' repeats Table~\ref{tab:main}. Last column: paired frame bootstrap 95\% interval, without ZB. (SFT), VLM and Context as in \cref{tab:main}.}
\label{tab:nodev}
\fitwidth{\begin{tabular}{lrrrrrrrrl}
\toprule
System & All 8 & w/o ZB & ALO7 w/o ZB & 5\,px & 6\,px & 7\,px & LA & \model $-$ it \\
\midrule
\model & \textbf{84.2} & \textbf{86.2} & \textbf{66.2} & \textbf{78.0} & \textbf{88.1} & \textbf{91.8} & \textbf{50.9} & -- \\
\quad w/o video context & \underline{82.6} & \underline{84.4} & \underline{62.4} & \underline{75.0} & \underline{86.3} & \underline{91.0} & \underline{47.9} & $+1.8$ \scriptsize[+1.4, +2.2] \\
\quad classifier only & 73.8 & 74.9 & 51.9 & 59.2 & 77.7 & 86.4 & 29.3 & $+11.3$ \scriptsize[+10.8, +11.8] \\
\midrule
Qwen3.5-9B (SFT) & 72.9 & 73.6 & 56.9 & 52.3 & 78.3 & 88.3 & 40.0 & $+12.6$ \scriptsize[+11.7, +13.5] \\
PP-OCRv6 (SFT) & 62.5 & 65.4 & 38.0 & 45.9 & 68.4 & 80.2 & 13.7 & $+20.8$ \scriptsize[+19.8, +21.8] \\
\midrule
Qwen3.5-9B (SFT) + LLM + Context & 62.4 & 69.3 & 52.6 & 47.7 & 73.5 & 84.7 & 36.3 & $+16.9$ \scriptsize[+15.7, +18.1] \\
PP-OCRv6 + VLM & 64.9 & 68.8 & 43.4 & 46.7 & 74.5 & 83.3 & 25.7 & $+17.4$ \scriptsize[+16.2, +18.6] \\
PP-OCRv6 + VLM + Context & 61.4 & 68.3 & 40.8 & 45.7 & 74.5 & 82.7 & 27.7 & $+17.9$ \scriptsize[+16.5, +19.3] \\
PP-OCRv6 (SFT) + VLM + Context & 70.0 & 73.9 & 49.2 & 54.6 & 78.9 & 86.6 & 31.6 & $+12.3$ \scriptsize[+11.2, +13.2] \\
\midrule
Qwen3.5-9B~\cite{qwen2026qwen35} & 65.9 & 70.3 & 44.2 & 50.2 & 75.1 & 83.9 & 27.0 & $+15.9$ \scriptsize[+14.6, +17.1] \\
Qwen3-VL-8B~\cite{bai2025qwen3vl} & 63.0 & 67.8 & 42.4 & 46.6 & 72.1 & 82.8 & 25.4 & $+18.4$ \scriptsize[+17.0, +19.8] \\
Qwen3-VL-32B~\cite{bai2025qwen3vl} & 57.2 & 61.7 & 33.2 & 37.9 & 66.9 & 78.1 & 21.6 & $+24.5$ \scriptsize[+22.6, +26.5] \\
HunyuanOCR (1B)~\cite{hunyuan2025ocr} & 54.3 & 58.7 & 31.0 & 36.4 & 61.3 & 76.4 & 13.6 & $+27.5$ \scriptsize[+26.2, +28.9] \\
PP-OCRv6 medium~\cite{zhang2026ppocrv6} & 54.1 & 57.2 & 32.2 & 32.5 & 61.3 & 75.6 & 10.6 & $+29.0$ \scriptsize[+27.8, +30.2] \\
PaddleOCR-VL-1.6 (0.9B)~\cite{cui2025paddleocrvl} & 53.8 & 57.4 & 34.3 & 33.6 & 60.3 & 76.3 & 13.2 & $+28.8$ \scriptsize[+27.6, +30.0] \\
MiniCPM-OCR 1.0~\cite{openbmb2026minicpmocr} & 52.8 & 56.1 & 34.7 & 33.9 & 58.4 & 74.0 & 13.7 & $+30.1$ \scriptsize[+29.1, +31.1] \\
GLM-OCR (0.9B)~\cite{duan2026glmocr} & 52.1 & 55.9 & 28.3 & 31.8 & 59.1 & 74.6 & 12.2 & $+30.3$ \scriptsize[+29.0, +31.6] \\
\bottomrule
\end{tabular}}
\end{table*}

\begin{table*}[t]
\centering
\small
\caption{Chinese characters read correctly (\%) by stroke count (Unihan kTotalStrokes), 5--7\,px pooled.}
\label{tab:complexity}
\fitwidth{\begin{tabular}{lrrrrr}
\toprule
System & 1--5 & 6--8 & 9--11 & 12--14 & $\geq$15 \\
\midrule
\textit{Game sub-benches} &  &  &  &  &  \\
\model & \textbf{88.7} & \textbf{87.4} & \textbf{79.3} & \textbf{80.4} & \textbf{68.0} \\
Qwen3.5-9B (SFT) & \underline{82.0} & \underline{77.6} & \underline{66.2} & \underline{65.9} & \underline{52.3} \\
Qwen3.5-9B~\cite{qwen2026qwen35} & 75.7 & 73.0 & 55.1 & 59.7 & 50.0 \\
PP-OCRv6 medium~\cite{zhang2026ppocrv6} & 66.7 & 58.6 & 41.3 & 42.2 & 30.2 \\
\quad\textit{characters} & 13,333 & 19,838 & 13,789 & 4,614 & 662 \\
\midrule
\textit{BCTR + OmniDocBench} &  &  &  &  &  \\
\model & 83.4 & \underline{82.3} & \underline{77.2} & \underline{71.7} & 68.4 \\
Qwen3.5-9B (SFT) & \underline{84.9} & 82.2 & 76.3 & 71.0 & \underline{70.0} \\
Qwen3.5-9B~\cite{qwen2026qwen35} & \textbf{85.2} & \textbf{82.7} & \textbf{77.3} & \textbf{73.0} & \textbf{72.1} \\
PP-OCRv6 medium~\cite{zhang2026ppocrv6} & 81.5 & 75.7 & 68.5 & 63.6 & 60.4 \\
\quad\textit{characters} & 54,084 & 81,792 & 47,205 & 18,291 & 4,093 \\
\midrule
\textit{Real-CE (5--7\,px)} &  &  &  &  &  \\
\model & 62.0 & \underline{55.0} & \underline{47.9} & \underline{45.0} & 41.5 \\
Qwen3.5-9B (SFT) & \underline{63.0} & 53.8 & 45.4 & 42.0 & \underline{41.9} \\
Qwen3.5-9B~\cite{qwen2026qwen35} & \textbf{64.0} & \textbf{56.4} & \textbf{49.6} & \textbf{47.9} & \textbf{44.6} \\
PP-OCRv6 medium~\cite{zhang2026ppocrv6} & 55.5 & 43.9 & 35.1 & 31.4 & 24.5 \\
\quad\textit{characters} & 11,805 & 17,889 & 11,886 & 5,112 & 984 \\
\bottomrule
\end{tabular}}
\end{table*}

\begin{table*}[t]
\centering
\small
\caption{How ambiguous is a tiny Chinese character? The 3{,}500 Chinese classes of our codebook rendered in four fonts and degraded like the bench (sub-pixel phase, area downsampling, JPEG q75), with no learning. Ranges are over the fonts (Noto Sans SC regular and bold, Resource Han Rounded, Noto Serif SC); the last row is the mean over the 12 template / query font pairs.}
\label{tab:ambiguity}
\fitwidth{\begin{tabular}{lrrrrrrrrr}
\toprule
\% of the 3{,}500 classes & 4\,px & 5\,px & 6\,px & 7\,px & 8\,px & 10\,px & 12\,px & 16\,px & 24\,px \\
\midrule
Characters with a look-alike & 97--98 & 90--91 & 70--78 & 51--56 & 39--40 & 19--22 & 8--13 & 2--3 & 0--0 \\
Template matching, font known (top-1) & 10--14 & 28--35 & 48--55 & 64--71 & 74--80 & 84--90 & 90--96 & 97--99 & 100--100 \\
Template matching, font known (top-5) & 18--23 & 41--48 & 62--68 & 76--82 & 85--89 & 92--97 & 96--100 & 100--100 & 100--100 \\
Template matching, another font (top-1) & 6 & 21 & 37 & 54 & 65 & 78 & 85 & 93 & 96 \\
\bottomrule
\end{tabular}}
\end{table*}

\begin{table*}[t]
\centering
\small
\caption{Cost per line on one A40 GPU (batched; model loading excluded) on the game sub-benches. \model: \lpd in rounds 1 and 2, each decoding two prompt modes (an upper bound). (SFT), VLM and Context as in \cref{tab:main}.}
\label{tab:efficiency}
\fitwidth{\begin{tabular}{llrr}
\toprule
System & Parameters & s / line & Chinese AR \\
\midrule
Qwen3.5-9B~\cite{qwen2026qwen35} & 9B & 0.142 & 65.9 \\
Qwen3.5-9B (SFT) & 9B & 0.123 & \underline{72.9} \\
Qwen3-VL-8B~\cite{bai2025qwen3vl} & 8B & 0.111 & 63.0 \\
Qwen3-VL-32B~\cite{bai2025qwen3vl} & 32B & 0.262 & 57.2 \\
HunyuanOCR (1B)~\cite{hunyuan2025ocr} & 1B & 0.046 & 54.3 \\
PaddleOCR-VL-1.6 (0.9B)~\cite{cui2025paddleocrvl} & 0.9B & 0.077 & 53.8 \\
GLM-OCR (0.9B)~\cite{duan2026glmocr} & 0.9B & 0.020 & 52.1 \\
MiniCPM-OCR 1.0~\cite{openbmb2026minicpmocr} & -- & 0.107 & 52.8 \\
PP-OCRv6 medium~\cite{zhang2026ppocrv6} & $<$0.1B & 0.004 & 54.1 \\
PP-OCRv6 (SFT) & $<$0.1B & 0.004 & 62.5 \\
PP-OCRv5 server~\cite{cui2025paddleocr3} & $<$0.1B & 0.007 & 49.3 \\
PP-OCRv6 + simple-csc & 7B + $<$0.1B & 0.445 & 54.2 \\
PP-OCRv6 (SFT) + VLM + Context & 9B + $<$0.1B & 0.190 & 70.0 \\
\model & 9B LM + small CNNs & 0.691 & \textbf{84.2} \\
\bottomrule
\end{tabular}}
\end{table*}

\section{Whole ALO7 frames}
\label{app:frames}

\begin{figure*}[t]
  \centering
  \includegraphics[width=\textwidth]{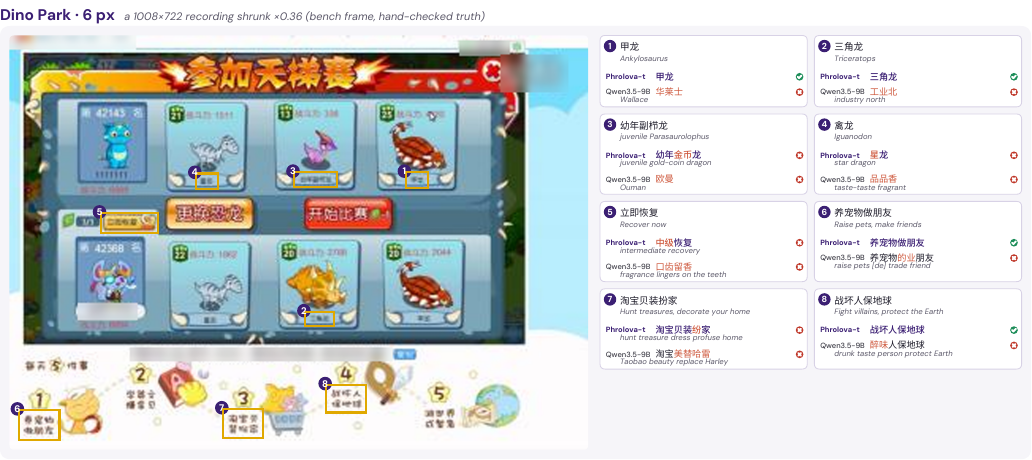}
  \caption{\textbf{A Dino Park frame at 6\,px.} \model reads the dinosaur names that Qwen3.5-9B turns into
  \gl{Wallace} and \gl{industry north}, but misses two rarer species names, a button label and one character of a slogan. Game frames:
  research use only, no commercial use.}
  \label{fig:alo7dp}
\end{figure*}

\begin{figure*}[t]
  \centering
  \includegraphics[width=\textwidth]{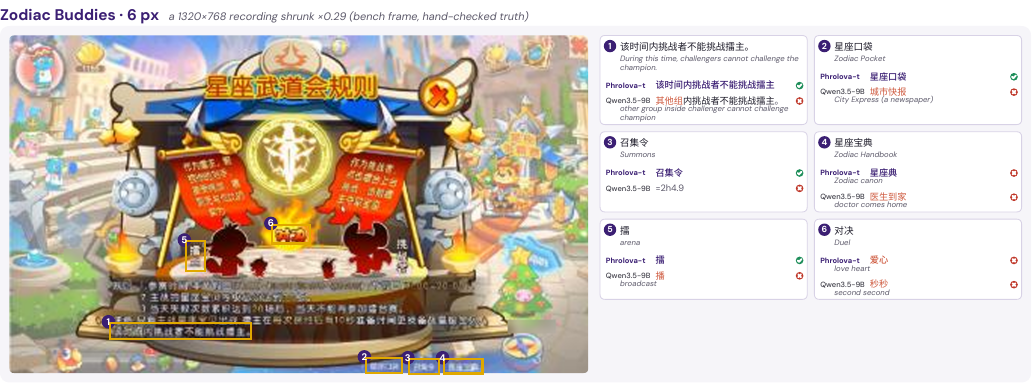}
  \caption{\textbf{A Zodiac Buddies frame at 6\,px} (the rules screen of the arena). Game frames: research use only,
  no commercial use.}
  \label{fig:alo7zb}
\end{figure*}

\begin{figure*}[t]
  \centering
  \includegraphics[width=\textwidth]{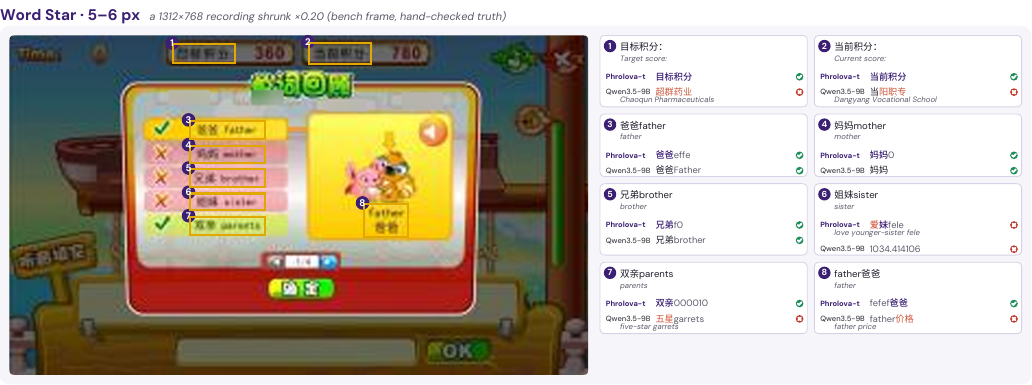}
  \caption{\textbf{A Word Star frame at 5--6\,px} (a vocabulary review). The Latin letters are not scored. Game
  frames: research use only, no commercial use.}
  \label{fig:alo7ws}
\end{figure*}

\cref{fig:alo7dp,fig:alo7zb,fig:alo7ws} show one whole benchmark frame each of Dino Park, Zodiac Buddies and Word Star at the resolution the readers
get, shrunk from the native recording exactly as the benchmark set was built. The circled lines are targets; each card
gives the hand-checked truth with its English, and the readings of \model and Qwen3.5-9B (coral: wrong Chinese
characters; grey: literal English of a wrong reading; only Chinese characters are scored, so Latin letters and digits
are shown muted). We show successes and failures of \model alike. Player names and the account number are blurred.

\section{Failure cases and limitations}
\label{app:failures}

\noindent\textbf{Context can mislead.} The document prompt carries the game's vocabulary, and the language model sometimes
prefers it to what the pixels say: on Zodiac Buddies, \gl{Buddy training camp} is read as
\gl{my Zodiac street}, a phrase from elsewhere in the game.

\noindent\textbf{Context gating.} \gbank assumes that a document has other text. With none, as in BCTR and Real-CE, it costs
1.1--2.3 points on BCTR and 3.0 at 5\,px on Real-CE. Running the glyph round only when a document has other text, a
rule we did not use for the reported results, gives 80.0\% on BCTR and OmniDocBench, a statistical tie with the best
system ($-0.5$ \scriptsize[$-2.0$, $+0.3$]\normalsize).

\noindent\textbf{Closed vocabulary and speed.} \clsnet has 3,548 classes. The 304 benchmark characters outside them (0.6\%)
are never read, where the VLMs read 31--41\% of them. Reading takes 0.69\,s per line, against about 0.13\,s for a 9B VLM.

\noindent\textbf{Ground truth.} Five sub-benches (Word Star and the four miHoYo sub-benches) were drafted by PP-OCR at native resolution and checked by hand. This could
favour readers that err like PP-OCR, but every reader, including PP-OCR itself, reads 98--99\% at native resolution,
so drafting errors are rare compared with the differences we report.

\end{document}